\documentclass{article}

\PassOptionsToPackage{numbers}{natbib}
\usepackage[preprint]{neurips_2026}

\usepackage[utf8]{inputenc}
\usepackage[T1]{fontenc}
\usepackage{hyperref}
\usepackage{makecell}
\usepackage{url}
\usepackage{booktabs}
\usepackage{amsfonts}
\usepackage{amsmath}
\usepackage{amssymb}
\usepackage{nicefrac}
\usepackage{microtype}
\usepackage{xcolor}
\usepackage{graphicx}
\usepackage{multirow}
\usepackage{enumitem}
\usepackage{placeins}
\usepackage{wrapfig}
\usepackage[most]{tcolorbox}
\tcbuselibrary{listings}
\usepackage{graphicx}
\usepackage{subcaption} 
\usepackage{tabularx}
\usepackage{booktabs}
\usepackage{makecell}
\usepackage{amssymb} 

\graphicspath{{figs/}}

\definecolor{promptblue}{HTML}{2B5B9E}
\definecolor{prompttea}{HTML}{1F7A6F}
\definecolor{promptplum}{HTML}{6B4C9A}
\definecolor{promptamber}{HTML}{B45A1C}
\definecolor{promptolive}{HTML}{556B2F}

\newtcolorbox{promptbox}[2][promptblue]{
  enhanced,
  colback       = #1!4!white,
  colframe      = #1!80!black,
  colbacktitle  = #1!85!black,
  coltitle      = white,
  fonttitle     = \bfseries\small,
  title         = {#2},
  boxrule       = 0.6pt,
  titlerule     = 0pt,
  arc           = 3pt,
  left=8pt, right=8pt, top=6pt, bottom=6pt,
  fontupper     = \small,
}

\lstdefinestyle{jsontool}{
  basicstyle        = \scriptsize\ttfamily,
  breaklines        = true,
  breakatwhitespace = true,
  showstringspaces  = false,
  columns           = fullflexible,
  keepspaces        = true,
  stringstyle       = \color{promptplum!85!black},
  keywordstyle      = \color{promptblue!85!black}\bfseries,
  commentstyle      = \color{gray!70!black}\itshape,
  morestring        = [b]",
  morecomment       = [l]{//},
  morekeywords      = {true,false,null},
  literate          = {:}{{\textcolor{black}{:}}}1,
}
\newtcblisting{toolsjson}[2][promptolive]{
  enhanced,
  colback       = #1!4!white,
  colframe      = #1!80!black,
  colbacktitle  = #1!85!black,
  coltitle      = white,
  fonttitle     = \bfseries\small,
  title         = {#2},
  boxrule       = 0.6pt,
  titlerule     = 0pt,
  arc           = 3pt,
  left=8pt, right=8pt, top=6pt, bottom=6pt,
  listing only,
  listing options = {style=jsontool},
}

\title{I-WebGenBench : Evaluating Interactivity in LLM-Generated Scientific Web Applications}

\author{
Dasen Dai$^{1*}$,
Biao Wu$^{2*}$,
Meng Fang$^{3}$,  
Shuoqi Li$^{1}$,
Wenhao Wang$^{1\dagger}$ \\
$^{1}$Vast Intelligence Lab, $^{2}$UTS, $^{3}$University of Liverpool \\
$^{*}$Equal contribution, $^{\dagger}$Corresponding author
}

\begin{document}

\maketitle

\begin{abstract}
Interactive scientific concepts such as optimization dynamics, chemical processes, and economic equilibria are inherently dynamic and require user-driven exploration for effective understanding. Recent advances in LLMs have made it possible to generate complex web applications from natural language, raising the prospect of automatically creating interactive scientific demonstrations from research papers. However, existing evaluation frameworks for LLM-based code generation focus primarily on static correctness or visual fidelity, and fail to assess whether generated interfaces exhibit meaningful, event-driven interactivity. We introduce I-WebGenBench, the first benchmark for evaluating the ability of LLMs to generate interactive scientific web applications from paper-derived specifications. The benchmark consists of 201 curated tasks spanning five scientific domains, each requiring models to construct a complete application from scratch while preserving both scientific fidelity and interaction requirements. We propose a deterministic Interaction Probe that executes semantic user actions and detects DOM mutations as a signal of responsiveness, enabling a dual-metric evaluation with Build Success Rate and Interaction Rate that explicitly decouples compilability from interactivity. Experiments on a diverse set of models reveal a substantial gap between compilation and interaction: while most systems achieve high BSR, their IR remains significantly lower, indicating limited ability to implement functional, event-driven behavior. Further analysis shows that this gap is primarily driven by deficiencies in interaction logic rather than rendering quality, and is strongly correlated with structural properties such as state management complexity and layout design. These findings highlight a fundamental limitation of current LLMs in translating structural generation into interactive behavior, and establish I-WebGenBench as a new testbed for advancing interactive code generation.
\end{abstract}

\section{Introduction}

Scientific concepts such as gradient descent, chemical kinetics, and Nash equilibria are inherently dynamic and parameter-sensitive. However, they are typically presented in static formats such as PDFs, which limits opportunities for interactive exploration~\cite{chen2025paper2web,dai2026papervoyager,lu2025webgen,chen2023webvln,yao2022webshop,xie2025swefixertrainingopensourcellms}. Recent advances in large language models (LLMs) have significantly improved their ability to generate code, and when combined with agent frameworks, they can now produce increasingly complex applications from natural language descriptions~\cite{zhang2023repocoder,zhuo2024bigcodebench,jain2024livecodebench,jiangwebgen,yao2022webshop,shi2025presentagent,dong2025agenticreinforcedpolicyoptimization,yang2024sweagent,lu2025uxagent,xia2024agentless}. This progress makes it plausible to automatically generate interactive scientific demonstrations that enable real-time, hands-on understanding of abstract concepts.

However, existing evaluation frameworks for LLM-based code generation do not adequately capture interactive capabilities. Code benchmarks such as HumanEval~\cite{chen2021evaluating} and SWE-bench~\cite{yang2024swe,jimenez2023swe,aleithan2024swe} focus on algorithmic correctness of isolated programs, ignoring whether generated interfaces respond to user inputs. Front-end design-to-code benchmarks, including Design2Code~\cite{si2025design2codebenchmarkingmultimodalcode}, Interaction2Code~\cite{xiao2025interaction2code}, and WebVR~\cite{barsoum2005webvr}, primarily assess static visual reconstruction without systematically evaluating interactive behavior. More recent systems such as PaperVoyager~\cite{dai2026papervoyager} attempt to convert research papers into interactive applications, but are limited in scale and do not distinguish between compilability and genuine interactivity. Similarly, WebCompass~\cite{lei2026webcompass} incorporates interaction evaluation via browser agents, yet lacks a deterministic, DOM-mutation-based metric to decouple execution from responsiveness, and does not analyze structural factors such as state management and layout design. As a result, there is currently no systematic and reliable method to evaluate whether generated scientific web applications exhibit meaningful, event-driven interactivity across diverse domains.

To this end, we introduce I-WebGenBench, a benchmark designed to systematically evaluate the ability of LLM-based agents to generate interactive scientific web applications from paper-derived specifications. Unlike prior code-generation benchmarks that focus on functional correctness or patching existing codebases, our benchmark requires models to construct complete, interactive applications from scratch, capturing both scientific fidelity and event-driven behavior~\cite{liu2023repobench,liu2026webcoderbench,zhuo2024bigcodebench,xu2025webbenchllmcodebenchmark,yang2024swe,guo2024iwbenchevaluatinglargemultimodal}. Two key challenges arise in building such a benchmark: (1) how to curate diverse and scientifically grounded specifications that reflect real-world interactive concepts, and (2) how to reliably evaluate whether generated applications exhibit genuine interactivity beyond successful compilation. I-WebGenBench addresses these challenges through a structured data construction process and a deterministic evaluation pipeline.

To construct the dataset, we first collect a diverse set of research papers across five domains—Biology, Chemistry, Computer Science, Economics, and Human-Computer Interaction (HCI). For each paper, domain experts identify core mechanisms suitable for interactive exploration and translate them into structured specifications, including required UI components, controllable parameters, and expected behavioral responses. These specifications are then standardized into natural language prompts that define interaction requirements without prescribing implementation details. To ensure quality and consistency, all specifications are manually reviewed and refined by multiple annotators, with disagreements resolved through discussion. As shown in Tabel ~\ref{tab:benchmark_comparison}, This process yields a corpus of 201 high-quality specifications with varying abstraction levels and interaction complexity. For evaluation, we design an automated Interaction Probe that executes semantically valid user actions on generated applications and detects DOM mutations as a binary signal of responsiveness. Combined with Build Success Rate (BSR), which measures compilability, and Interaction Rate (IR), which measures reactivity, our framework provides a systematic and fine-grained assessment of both execution and interaction behavior.

Using this evaluation framework, we benchmark a diverse set of representative models, including proprietary LLMs, open-source code models, and an API-based agent, on I-WebGenBench.    Our results reveal a consistent and substantial gap between compilability and interactivity: while most models achieve high BSR, indicating reliable code generation and deployment, their IR remains significantly lower, demonstrating limited ability to implement functional, event-driven behaviors. Further analysis shows that this gap is primarily driven by deficiencies in interaction logic rather than visual rendering or content alignment. In particular, models often generate structurally complete interfaces with numerous interactive elements, but only a fraction of these respond to user actions. We also identify strong correlations between interaction success and structural properties of the generated code, including state management complexity, layout patterns , and design choices such as visual themes. These findings highlight that achieving true interactivity requires not only syntactic correctness, but also coherent stateful execution, which remains a key limitation of current models.

\begin{table}[t!]
\centering
\small
\caption{Comparison of I-WebGen-Bench with other repository-level software engineering benchmarks. 
We follow the statistical protocol of~\cite{lu2025webgen} for consistency with prior work.}
\vspace{2mm}
\resizebox{0.8\columnwidth}{!}{%
\begin{tabular}{lcccc}
\toprule
\textbf{Benchmark} & \textbf{From Scratch} & \textbf{Number of Files} & \textbf{Lines of Code} & \textbf{Interactive} \\
\midrule

SWE-Bench~\cite{jimenez2023swe}              & $\times$ & 1.7  & 32.8  & $\times$ \\
SWE-Bench Multimodal$^*$~\cite{yang2024swe} & $\times$ & 2    & 27    & $\times$ \\
SWE-Lancer~\cite{miserendino2025swe}        & $\times$ & 2    & 55    & $\times$ \\
WebGenBench~\cite{lu2025webgen}             & $\checkmark$ & 8.1  & 315.3 & $\times$ \\
PaperVoyager~\cite{dai2026papervoyager}     & $\checkmark$ & 11.9 & 698   & $\times$ \\
\midrule
I-WebGenBench                                & $\checkmark$ & 21.2 & 951   & $\checkmark$ \\
\bottomrule
\end{tabular}%
}
\label{tab:benchmark_comparison}
\vspace{-5mm}
\end{table}

We summarize our main contributions as follows:
\begin{itemize} 
\item We introduce I-WebGenBench, the first benchmark for evaluating LLM-generated \emph{interactive scientific web applications} from paper-derived specifications. It comprises 201 curated tasks across five scientific domains with diverse abstraction levels, requiring models to build complete applications from scratch while preserving scientific fidelity and interaction requirements.
\item We propose Interaction Probe that executes semantic user actions and detects DOM mutations to measure interactivity. Combined with BSR and IR, our framework explicitly decouples compilability from true user responsiveness, enabling the first systematic and pixel-independent evaluation of interactive behavior.
\item We conduct a comprehensive empirical study revealing a persistent compilation--interaction gap in current models. Our analysis shows that interaction failures stem primarily from missing event-driven logic rather than rendering issues, and identifies key structural factors such as state management complexity and layout patterns that govern interactive success.
\end{itemize}
\begin{figure}[t!]
  \centering
  \includegraphics[width=1\columnwidth]{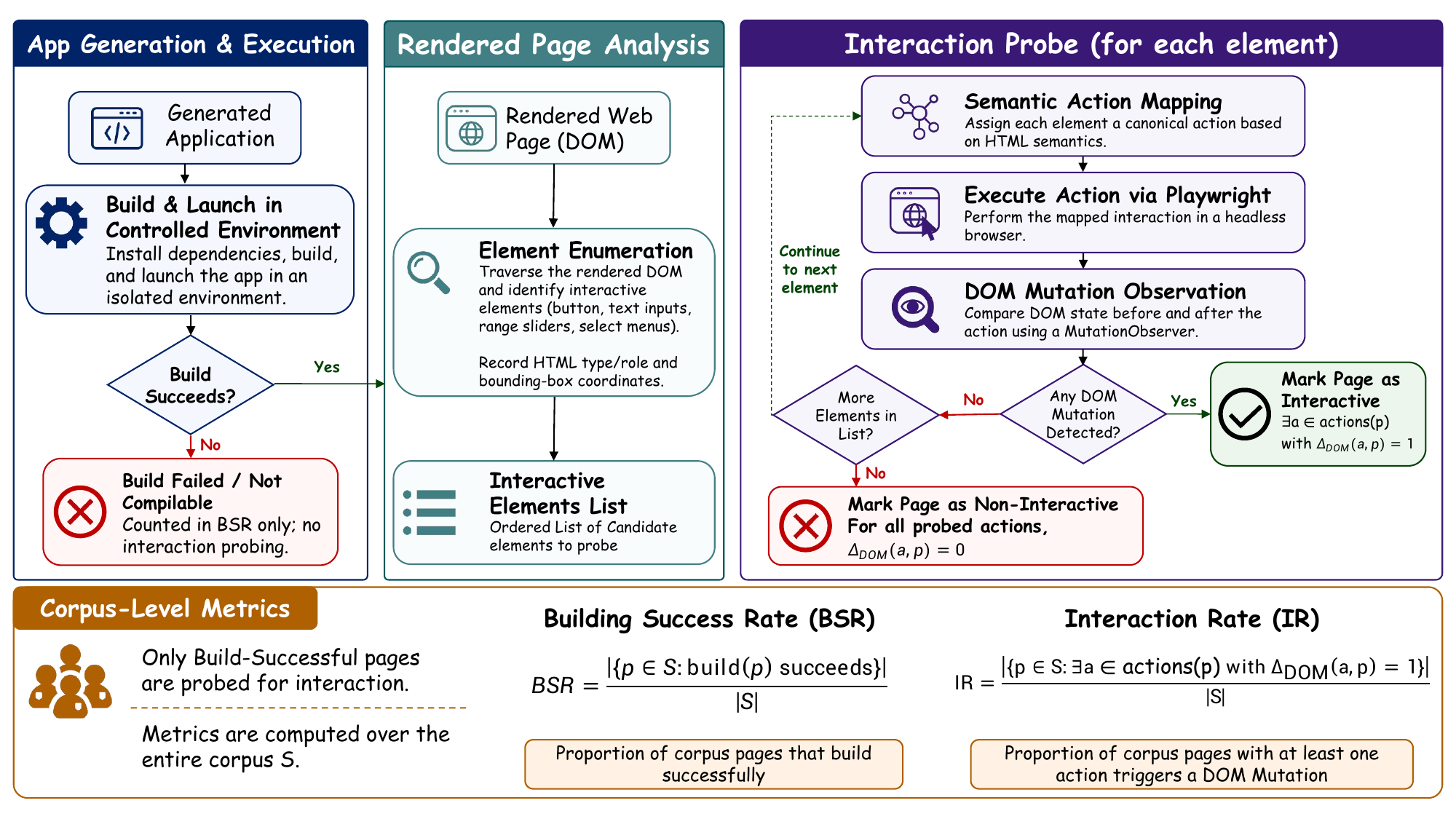}
  \caption{Overview of the I-WebGenBench evaluation pipeline. Generated applications are built in a controlled environment, followed by DOM-based analysis and interaction probing. Interactive elements are enumerated and executed via browser actions, and DOM mutations are used to determine responsiveness. BSR measures compilability, while I) measures whether user actions trigger observable UI changes.}
  \label{fig:benchmark_architecture}
\end{figure}

\section{Related Work}

\subsection{Benchmarks for Coding LLMs.}
 
Numerous benchmarks have been proposed to evaluate the coding capabilities of large language models. Early evaluations mainly focus on well-specified programming tasks, where models generate functions or short programs that satisfy given test cases~\cite{hendrycks2021measuring,chen2021evaluating,austin2021program}. These datasets are typically collected from natural-language programming queries~\cite{zhang2024naturalcodebench}, competitive programming tasks~\cite{jain2024livecodebench}, automatically synthesized problems~\cite{zhuo2024bigcodebench}, and expert-designed evaluations~\cite{muennighoff2023octopack}. While effective for measuring code correctness, these benchmarks mainly evaluate function-level generation on relatively small, self-contained programs. More recent benchmarks extend evaluation to repository-level tasks that better reflect real-world development. For example, SWE-bench~\cite{jimenez2023swe,yang2024swe} and SWE-Lancer~\cite{miserendino2025swe} are built from real software repositories and issue reports, requiring models to implement fixes or features within existing codebases. These tasks involve bug fixing~\cite{jimenez2023swe,yang2024swe,aleithan2024swe}, code completion~\cite{liu2023repobench,zhang2023repocoder}, and multi-file patch generation~\cite{miserendino2025swe}. 

\subsection{Web Generation Agents} 

Recent advances in LLMs have enabled systems that assist or automate software and web development. Agent-style frameworks such as OpenHands~\cite{wang2024openhands}, SWE-agent~\cite{yang2024sweagent}, and Aider~\cite{aiderai2024aider} allow models to interact with tools and execution environments to iteratively generate and refine code. Development assistants including Cursor~\cite{Cursor2024cursor}, GitHub Copilot~\cite{GitHub2024copilot}, and Devin~\cite{Wu2024devin} further demonstrate the potential of LLM-driven programming systems. Beyond general code generation, recent work explores integrating visual information into coding tasks. In particular, VLM have been studied for generating webpage code from visual inputs, such as reconstructing HTML from webpage screenshots~\cite{guo2024iwbenchevaluatinglargemultimodal,si2025design2codebenchmarkingmultimodalcode,yun2024web2codelargescalewebpagetocodedataset,beltramelli2017pix2codegeneratingcodegraphical}. Other studies evaluate models on implementing interactive elements or solving web development tasks through predefined pipelines~\cite{xiao2025interaction2codebenchmarkingmllmbasedinteractive,xiao2025designbenchcomprehensivebenchmarkmllmbased,xu2025webbenchllmcodebenchmark}. More recent work has begun to explore end-to-end web generation tasks. However, existing settings still primarily focus on static webpage generation. The problem of constructing dynamic web applications, as well as systematically evaluating their interaction behavior, remains underexplored~\cite{lu2025webgen,wang2025webgen,jiang2026webgen,lu2025webgen,liu2026webcoderbench,chen2025iwr}. In particular, there is a lack of benchmarks that assess whether generated interfaces support meaningful, event-driven interactions and how such capabilities vary across models.  In contrast, our work introduces a benchmark for dynamic webpage generation, focusing on the ability to construct interactive web systems from scratch and to analyze their interaction behavior.

\begin{figure}
    \centering
    \includegraphics[width=1\linewidth]{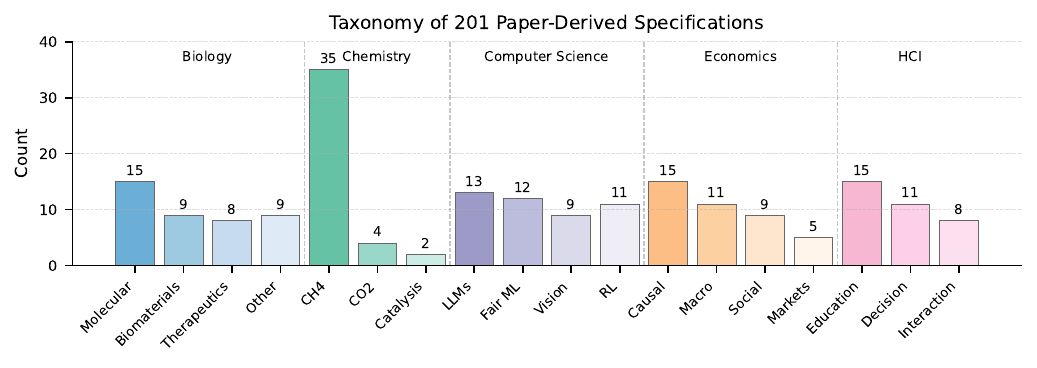}
    \vspace{-5mm}
    \caption{Taxonomy of the 201 paper-derived specifications across five domains.  Bars indicate the number of specifications per subcategory. The figure shows the distribution of topics within Biology, Chemistry, Computer Science, Economics, and HCI.}
    \label{fig:triple}
\end{figure}

\section{I-WebGenBench}



\subsection{Data Curation}

We construct I-WebGenBench from scientific paper PDFs spanning five domains: Biology, Chemistry, Computer Science, Economics, and Human--Computer Interaction (HCI). For each paper, the goal is to derive an interaction specification that defines the expected behaviors of an interactive scientific web application, including controllable parameters, dynamic visual components, and user-triggered responses.

To build a high-quality seed set, we recruit three PhD-level annotators for each domain. Papers are sampled from diverse research topics within each field, and annotators independently derive interaction specifications directly from the full PDFs. Each specification contains 3--10 core interaction elements describing parameter manipulation, dynamic updates, and expected system behavior. Across annotators, more than 95\% of identified interaction points are consistent, indicating strong agreement on the underlying interactive mechanisms. Disagreements are resolved through discussion to produce unified reference specifications.

To scale beyond fully manual annotation, we design a structured prompting pipeline that takes a paper PDF as input and automatically generates an interaction specification. The prompt guides the model to extract key scientific concepts, dynamic processes, and parameter-dependent behaviors, and to organize them into a standardized description of interactive elements. After calibration on the manually annotated seed set, the generated specifications achieve over 95\% overlap with human references. We therefore apply the same prompting pipeline to the remaining papers to efficiently expand dataset coverage across domains. Each generated specification is expressed as a natural-language prompt describing required UI components, controllable parameters, and expected system behaviors, without prescribing implementation details. Models are required to generate complete interactive applications solely from these specifications.

\paragraph{Quality Control.}
To ensure the reliability of the automatically generated specifications, we conduct an additional expert-review stage across all domains. For each domain, we randomly sample generated specifications and ask three domain experts to independently evaluate them against the source paper PDFs. Experts assess whether a specification accurately captures the essential interactive mechanisms of the paper, including controllable parameters, dynamic behaviors, and expected responses, while also checking for unsupported or spurious content. We adopt a majority-vote filtering protocol: a specification is retained only if at least two of the three experts judge it to be both consistent with the original paper and sufficiently complete for downstream evaluation. Specifications failing to satisfy this criterion are discarded. This additional review process reduces the risk of annotation noise and improves consistency across domains. After filtering, the final dataset contains 201 high-quality specifications spanning five scientific domains, which serve as standardized task definitions for evaluating generated interactive applications.

\subsection{Evaluation Methods}

We evaluate generated applications using two complementary paradigms: rule-based metrics and model-based assessment.

\paragraph{Rule-Based Metrics.}
We first assess execution validity and interaction behavior using deterministic metrics. 
A generated application is considered valid if it compiles successfully in a controlled environment, yielding the Build Success Rate (BSR):

\vspace{-3mm}
\begin{equation}
BSR = \frac{|\{p \in \mathcal{S}: \text{build}(p) \text{ succeeds}\}|}{|\mathcal{S}|}.
\end{equation}

To measure interaction, we define the Interaction Rate (IR) based on whether user actions produce observable effects:

\vspace{-7mm}
\begin{equation}
IR = \frac{|\{p \in \mathcal{S}: \exists\, a \in \text{actions}(p)\text{ with } \Delta_{DOM}(a,p) = 1\}|}{|\mathcal{S}|}.
\end{equation}

Here, $\Delta_{DOM}(a,p) = 1$ indicates that action $a$ triggers at least one DOM mutation. To compute IR, we use a deterministic interaction probe based on a headless browser (Playwright). The probe operates on the rendered DOM by identifying interactive elements, applying predefined actions, and detecting resulting DOM changes. Interaction is defined through DOM-level state changes rather than visual differences.

\paragraph{VLM-as-a-Judge.}
To assess properties not captured by rule-based metrics, we employ a VLM as a judge. For each compilable application, the VLM is provided with the interaction specification and a sequence of screenshots collected during interaction probing. Each application is scored on a 100-point scale across multiple dimensions, including semantic alignment, visual quality, interaction fidelity, clarity, and stability. Semantic alignment evaluates consistency with the specification, while visual quality assesses layout and interface design. Interaction fidelity examines whether user actions lead to meaningful updates, and clarity measures interpretability. Stability accounts for runtime errors and rendering failures. This model-based evaluation complements BSR and IR by capturing higher-level semantic and design properties that are not directly observable through programmatic signals. Further implementation details and evaluation protocols are described in the appendix.




\section{Experiments}

\subsection{Implementation Details}

All experiments prompt each large language model M to generate a complete web application A from a scientific specification S defined in a task instance T from the I-WebGenBench dataset. Each generated project is executed in a controlled browser-based environment that installs dependencies, builds the project, and launches it in a Chromium instance. The evaluation system uses Playwright to load the generated interface, execute interaction sequences, and capture runtime browser events. The Interaction Probe loads the application, invokes ExtractUI to identify interactive elements such as buttons, sliders, and input fields from the DOM, and then applies GenerateOps to construct candidate user operations including clicks, parameter adjustments, and text inputs. These operations are executed sequentially through Playwright while a MutationObserver monitors DOM mutation signals to determine whether interface responses occur after each action. For each task instance T, the pipeline records build status and logs interaction traces during automated exploration. All models are evaluated with identical prompts and execution environments, and the same interaction protocol is applied across generated applications to measure behavior under consistent conditions.

\paragraph{Models.}
Unlike prior work that relies on external code-agent frameworks, we directly evaluate the end-to-end capability of VLMs to generate interactive web applications from specifications~\cite{lu2025webgen}. All models are prompted to produce complete applications in a single pass without tool use or iterative refinement. The exact prompting protocol is provided in the appendix. We evaluate a diverse set of representative models, including proprietary API-based models, API-based agents, and open-source models. These include GPT-5.4~\cite{singh2025openai}, Gemini-3.1-pro-preview~\cite{comanici2025gemini}, Kimi-K2.5~\cite{team2025kimi}, Qwen-3.5~\cite{yang2025qwen3}, Qwen-3-Max~\cite{yang2025qwen3}, MiniMax-M2.7~\cite{li2025minimax}, DeepSeek-V4-Flash~\cite{guo2025deepseek}, MiMo-V2.5~\cite{xiaomi2025mimo}, Hunyuan-T1, and DeepSeek-V3.2~\cite{guo2025deepseek}, as well as the API-based agent PaperVoyager~\cite{dai2026papervoyager} and the open-source model WebGenAgent-SFT-8B~\cite{lu2025webgen}.

\begin{table*}[t!]
\centering
\caption{Overall scores, build success rate (BSR), and interaction rate (IR) across five domains. `$-$' indicates that IR is not reported because the model's build success rate is insufficient for meaningful interaction evaluation.}
\setlength{\tabcolsep}{3pt} 
\renewcommand{\arraystretch}{1.1}
\resizebox{\textwidth}{!}{
\begin{tabular}{lcccccccccccccccc}
\toprule
\multirow{2}{*}{\textbf{Model}}
& \multicolumn{3}{c}{\textbf{Bio}}
& \multicolumn{3}{c}{\textbf{Chem}}
& \multicolumn{3}{c}{\textbf{CS}}
& \multicolumn{3}{c}{\textbf{Econ}}
& \multicolumn{3}{c}{\textbf{HCI}}
& \textbf{AVG} \\

\cmidrule(lr){2-4}
\cmidrule(lr){5-7}
\cmidrule(lr){8-10}
\cmidrule(lr){11-13}
\cmidrule(lr){14-16}

& All & BSR & IR
& All & BSR & IR
& All & BSR & IR
& All & BSR & IR
& All & BSR & IR
& All \\

\midrule
\multicolumn{17}{l}{\textbf{Based on API-based Models}} \\

GPT-5.4
& 82.6 & 97.6 & 66.3
& 79.1 & 100.0 & 50.7
& 76.1 & 97.8 & 49.7
& 75.8 & 100.0 & 47.8
& 84.1 & 100.0 & 55.0
& 79.4 \\

Gemini-3.1
& 61.3 & 90.2 & 47.8
& 50.3 & 78.0 & 41.9
& 45.2 & 82.2 & 27.9
& 52.5 & 87.5 & 29.1
& 62.1 & 97.1 & 39.1
& 53.8 \\

Kimi-K2.5
& 46.7 & 100.0 & 51.9
& 54.3 & 100.0 & 30.5
& 51.8 & 100.0 & 44.1
& 38.9 & 97.5 & 34.1
& 52.8 & 100.0 & 47.5
& 48.9 \\

Qwen-3.5
& 55.7 & 100.0 & 13.4
& 58.1 & 100.0 & 15.6
& 41.2 & 93.3 & 19.2
& 43.2 & 100.0 & 15.1
& 38.3 & 94.3 & 12.0
& 47.5 \\

Qwen-3-Max
& 32.3 & 48.8 & 54.5
& 21.1 & 34.1 & 42.3
& 32.5 & 48.9 & 37.4
& 27.8 & 52.5 & 37.7
& 45.8 & 79.4 & 38.3
& 31.5 \\

MiniMax-M2.7
& 34.5 & 95.1 & 36.6
& 29.4 & 95.1 & 18.4
& 35.1 & 100.0 & 29.5
& 21.3 & 72.5 & 9.6
& 31.3 & 97.1 & 20.9
& 30.4 \\

DeepSeek-V4-Flash
& 17.5 & 85.4 & 38.3
& 17.2 & 90.2 & 21.6
& 21.6 & 82.2 & 26.6
& 22.3 & 85.0 & 22.1
& 25.7 & 85.3 & 30.6
& 20.7 \\

MiMo-V2.5
& 22.9 & 95.1 & 44.5
& 12.3 & 41.5 & 35.2
& 17.2 & 66.7 & 28.6
& 14.0 & 52.5 & 34.0
& 15.8 & 67.6 & 54.3
& 16.5 \\

Hunyuan-T1
& 7.3 & 36.6 & 36.5
& 11.1 & 36.6 & 33.8
& 16.6 & 53.3 & 32.7
& 12.3 & 45.0 & 26.5
& 18.6 & 73.5 & 26.2
& 13.0 \\

DeepSeek-V3.2
& 9.7 & 12.2 & $-$
& 9.5 & 12.2 & $-$
& 8.8 & 13.3 & $-$
& 11.3 & 20.0 & $-$
& 5.0 & 0.0 & $-$
& 9.0 \\

\midrule
\multicolumn{17}{l}{\textbf{Based on API Agents}} \\

PaperVoyager (Gemini-3.1)
& 76.3 & 97.6 & 56.3
& 56.6 & 75.6 & 39.5
& 58.8 & 91.1 & 35.5
& 66.0 & 97.5 & 31.7
& 74.1 & 100.0 & 38.3
& 65.9 \\

\midrule
\multicolumn{17}{l}{\textbf{Based on Open-Source LLM}} \\

WebGenAgent-SFT-8B
& 5.0 & 0.0 & $-$
& 5.0 & 0.0 & $-$
& 5.0 & 0.0 & $-$
& 5.0 & 0.0 & $-$
& 5.0 & 0.0 & $-$
& 5.0 \\

\bottomrule
\end{tabular}
}

\label{tab:merged_full}
\end{table*}

\subsection{Main Results}

Table~\ref{tab:merged_full} presents a unified evaluation of model performance across five domains by jointly considering overall scores and core capabilities, including BSR and IR. A clear hierarchy emerges: proprietary models consistently dominate across both aggregate performance and underlying capabilities, with GPT-5.4 achieving the best results in all domains, indicating strong end-to-end competence in generating visually coherent, semantically aligned, and functionally executable web applications. However, despite near-saturated BSR values (often above 95\%), IR remains substantially lower across all models, revealing a persistent gap between successful code generation and effective interactive behavior implementation. This discrepancy suggests that while most models can reliably produce syntactically correct and compilable interfaces, they struggle to translate user intent into consistent event-driven logic and state updates.

Across domains, performance differences further highlight task-specific challenges. Biology and HCI tasks generally yield higher scores and IR, whereas Computer Science and Economics exhibit consistently lower interaction performance, reflecting the increased difficulty of implementing structured reasoning, algorithmic logic, and precise state transitions in interactive settings. Mid-tier models such as Qwen-3.5 and MiniMax-M2.7 reinforce this pattern: although they achieve high BSR, their low IR indicates a tendency to generate visually complete but functionally shallow interfaces. In contrast, the API-based agent PaperVoyager demonstrates a more balanced profile, achieving competitive scores alongside relatively high IR, which underscores the effectiveness of system-level orchestration in bridging the gap between generation and interaction. At the lower end, models like DeepSeek-V3.2 and WebGenAgent-SFT-8B fail to achieve reliable build success, leading to negligible or unavailable interaction performance. Overall, the unified results expose a fundamental limitation of current approaches: robust code generation does not necessarily translate into meaningful interactivity, and closing this gap remains a central challenge for future web generation systems.


\paragraph{Gap Between Visual Quality and Interactivity.}
Table~\ref{tab:score_breakdown} provides a more fine-grained breakdown of model performance across different evaluation dimensions, including visual quality (V), interaction capability (I), topic alignment (T), clarity (C), and rule compliance (R). Across models, interaction scores exhibit the largest variation and remain substantially lower than visual scores in most cases, suggesting that implementing functional interactivity is consistently more challenging than generating visually plausible interfaces. Proprietary models such as GPT-5.4 achieve relatively balanced performance across all dimensions, while weaker open-source models show pronounced deficiencies in interaction capability and semantic alignment. In particular, several lower-performing models obtain moderate visual scores despite extremely limited interaction performance, indicating a tendency to prioritize static interface generation over executable behavior. Overall, these results further reinforce the gap between surface-level visual quality and meaningful interactive functionality.

\begin{table}[t!]
\centering
\caption{Detailed Score Distribution Across Models (V: Visual, I: Interaction, T: Topic, C: Clarity, R: Rule). All values are averaged over 201 task instances.}
\vspace{2mm}
\resizebox{\textwidth}{!}{ 
\begin{tabular}{l ccccc c c l ccccc c}
\toprule
\textbf{Model} & \textbf{V} & \textbf{I} & \textbf{T} & \textbf{C} & \textbf{R} & \textbf{Total} & & \textbf{Model} & \textbf{V} & \textbf{I} & \textbf{T} & \textbf{C} & \textbf{R} & \textbf{Total} \\
\midrule
GPT-5.4 & 24.5 & 27.8 & 12.9 & 9.1 & 5.0 & \textbf{79.4} & & PaperVoyager & 19.9 & 22.5 & 11.3 & 7.2 & 5.0 & \textbf{65.9} \\
Gemini-3.1-pro & 15.6 & 20.3 & 7.8 & 5.2 & 5.0 & \textbf{53.8} & & Kimi-K2.5 & 14.6 & 13.9 & 9.0 & 6.3 & 5.0 & \textbf{48.9} \\
Qwen-3.5 & 21.2 & 16.3 & 5.0 & 0.0 & 5.0 & \textbf{47.5} & & Qwen-3-Max & 7.5 & 10.5 & 5.5 & 3.5 & 4.5 & \textbf{31.5} \\
MiniMax-M2.7 & 9.2 & 6.6 & 5.8 & 3.9 & 5.0 & \textbf{30.4} & & DeepSeek-v4-Flash & 5.0 & 6.5 & 3.5 & 2.5 & 3.2 & \textbf{20.7} \\
MiMo-v2.5 & 5.5 & 4.5 & 2.5 & 1.5 & 2.5 & \textbf{16.5} & & Hunyuan-T1 & 4.0 & 3.5 & 2.0 & 1.5 & 2.0 & \textbf{13.0} \\
\bottomrule
\multicolumn{15}{l}{\footnotesize Max scores: V(30), I(40), T(15), C(10), R(5).}
\end{tabular}
} 
\label{tab:score_breakdown}
\vspace{-2mm}
\end{table}

\paragraph{BSR, IR, and Model Behavior.}

\begin{figure*}[t]
\centering
\includegraphics[width=0.4\textwidth]{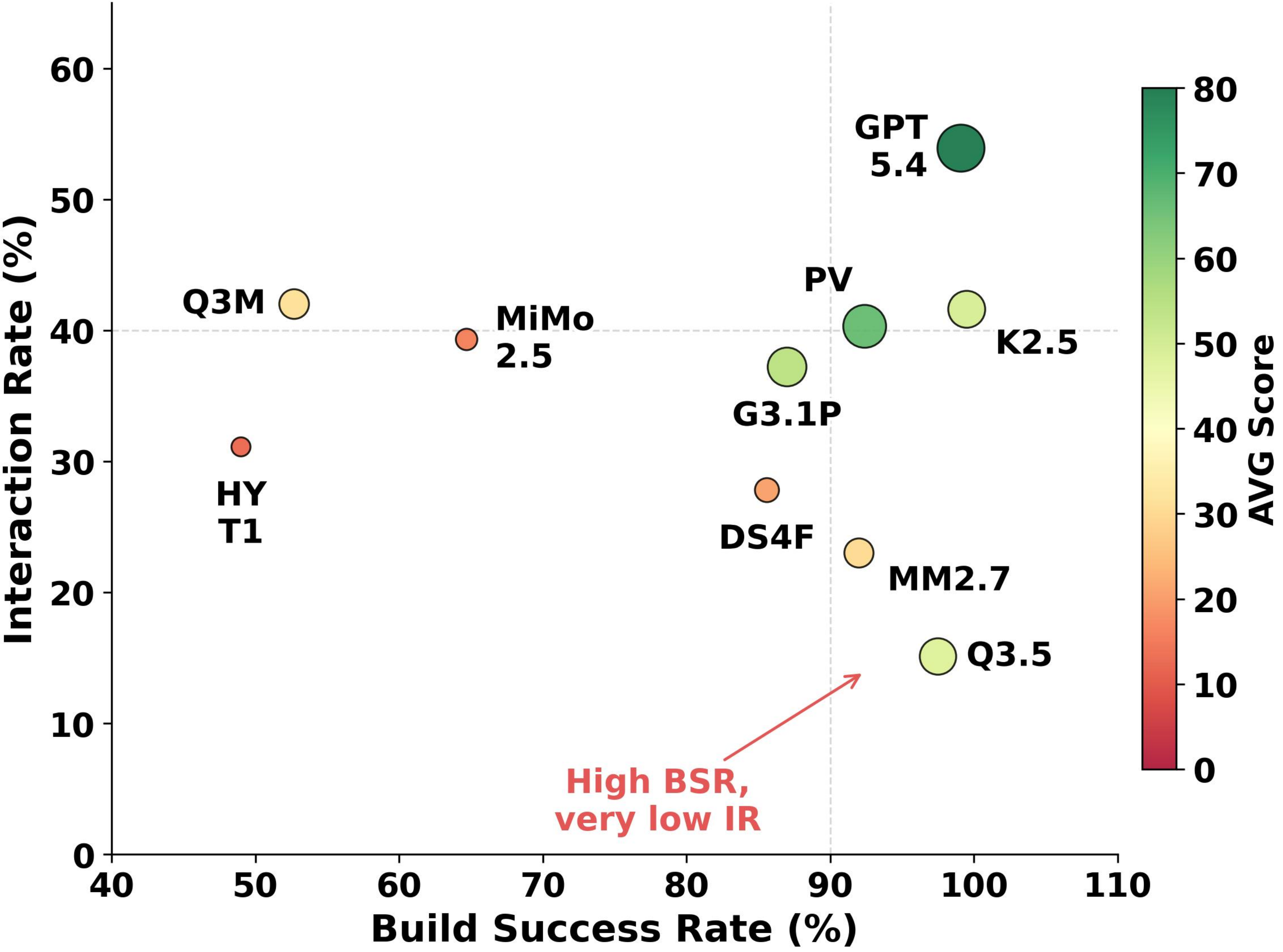}
\hfill
\includegraphics[width=0.55\textwidth]{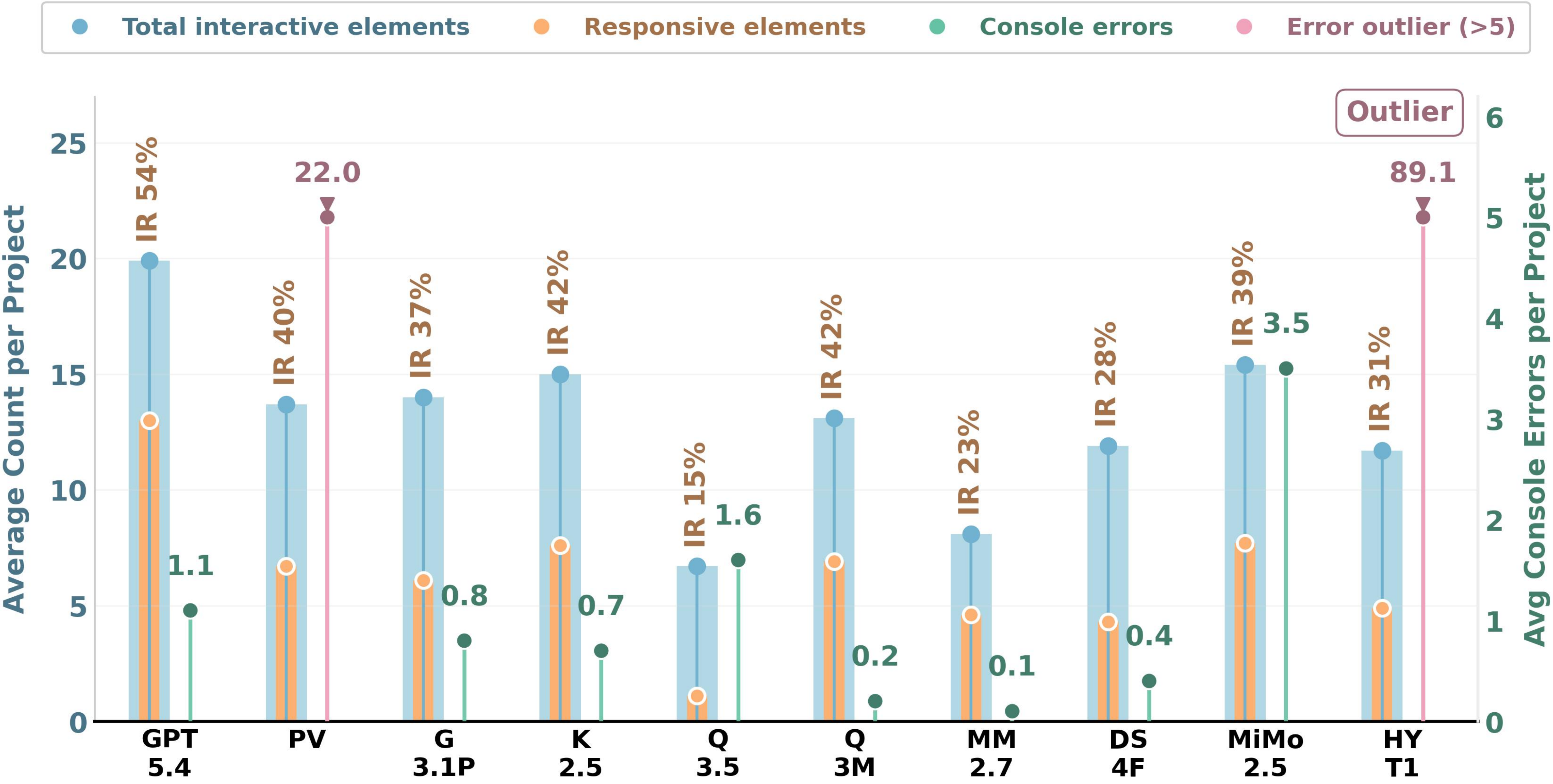}
\vspace{2mm}
\caption{(left) BSR vs IR. Bubble size and color indicate overall score. Most models achieve high BSR but substantially lower IR, confirming that functional interactivity remains the key bottleneck. (right) Interaction probe results: total interactive elements vs. responsive elements (left axis) and average console errors (right axis) per model; IR percentages are shown above each bar. Model abbreviations: GPT 5.4, PV, G3.1P, K2.5, Q3.5, Q3M, MM2.7, DS4F, MiMo 2.5, HYT1.}
\label{fig:two_pdfs}
\end{figure*}

In the left panel of Figure~\ref{fig:two_pdfs} , models exhibit distinct distributions in the two-dimensional space: most models cluster in regions with relatively high BSR (approximately 80\%–100\%), while IR values are more dispersed, primarily ranging from 10\% to 55\%. Some models (e.g., GPT-5.4 and PaperVoyager) appear in the upper-right region, characterized by both high BSR and relatively high IR. In contrast, other models (e.g., Qwen-3.5 and MiniMax-M2.7) show high BSR but comparatively low IR, forming a cluster in the lower-right region. Additionally, a small number of models (e.g., Qwen-3-Max) occupy regions with moderate BSR but relatively higher IR, exhibiting patterns that differ from the dominant distribution.

\paragraph{Effects of Generation Capacity.}
The right panel of Figure~\ref{fig:two_pdfs} summarizes model behavior under standardized generation settings. To minimize confounding factors, all models were evaluated using near-deterministic decoding (lowest supported temperature) and maximal output budgets, ensuring that observed differences primarily reflect intrinsic model capabilities rather than sampling variability. Under these conditions, a consistent pattern emerges: while most models can produce compilable code, their success is strongly constrained by generation capacity, including context length and available output budget. Models with restricted capacity exhibit systematically lower build success, indicating that insufficient sequence budget directly limits end-to-end generation performance. At the extreme, models operating under tight constraints fail to produce any valid outputs, highlighting the difficulty of completing complex generation tasks within limited context. These findings suggest that, beyond raw modeling ability, generation capacity—particularly sequence length and available output budget—is a critical bottleneck for real-world usability.

\begin{figure}[t]
  \centering
  \begin{minipage}[t]{0.48\columnwidth}\centering
    \includegraphics[width=\linewidth]{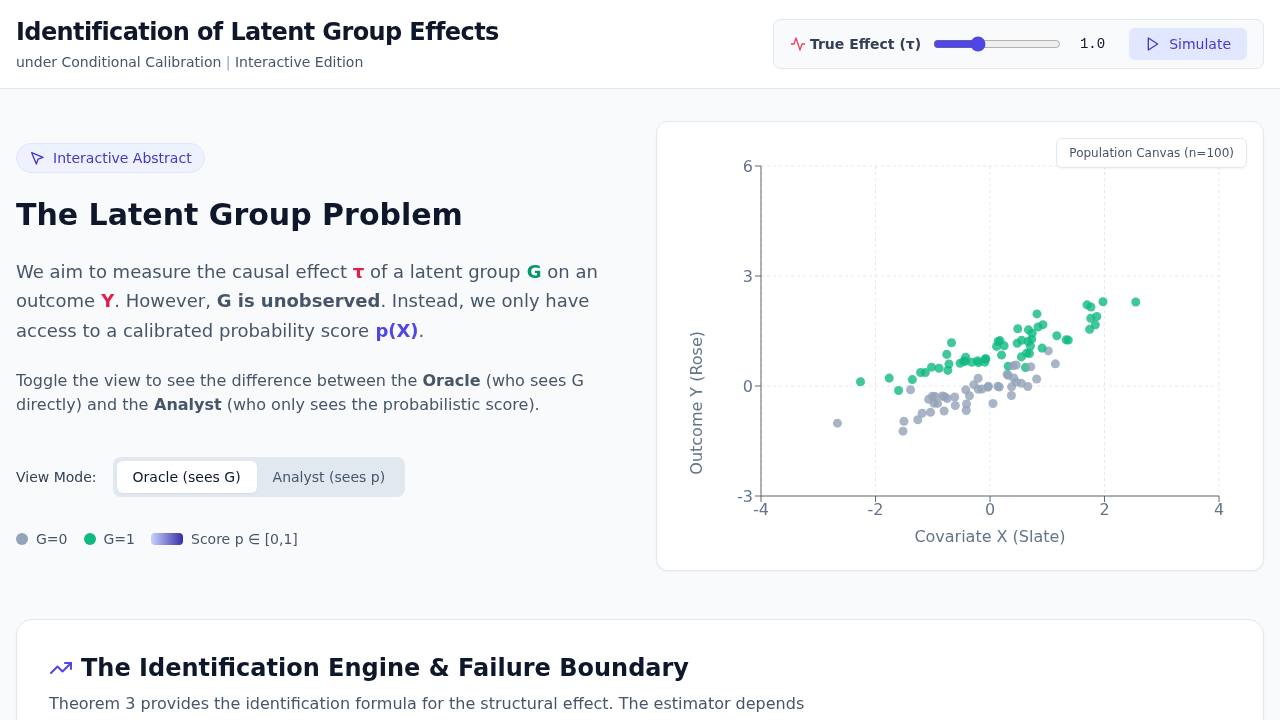}\\
    {\scriptsize (a) Kimi-K2.5: Visual 58/70, IR \textbf{0\%}}
  \end{minipage}\hfill
  \begin{minipage}[t]{0.48\columnwidth}\centering
    \includegraphics[width=\linewidth]{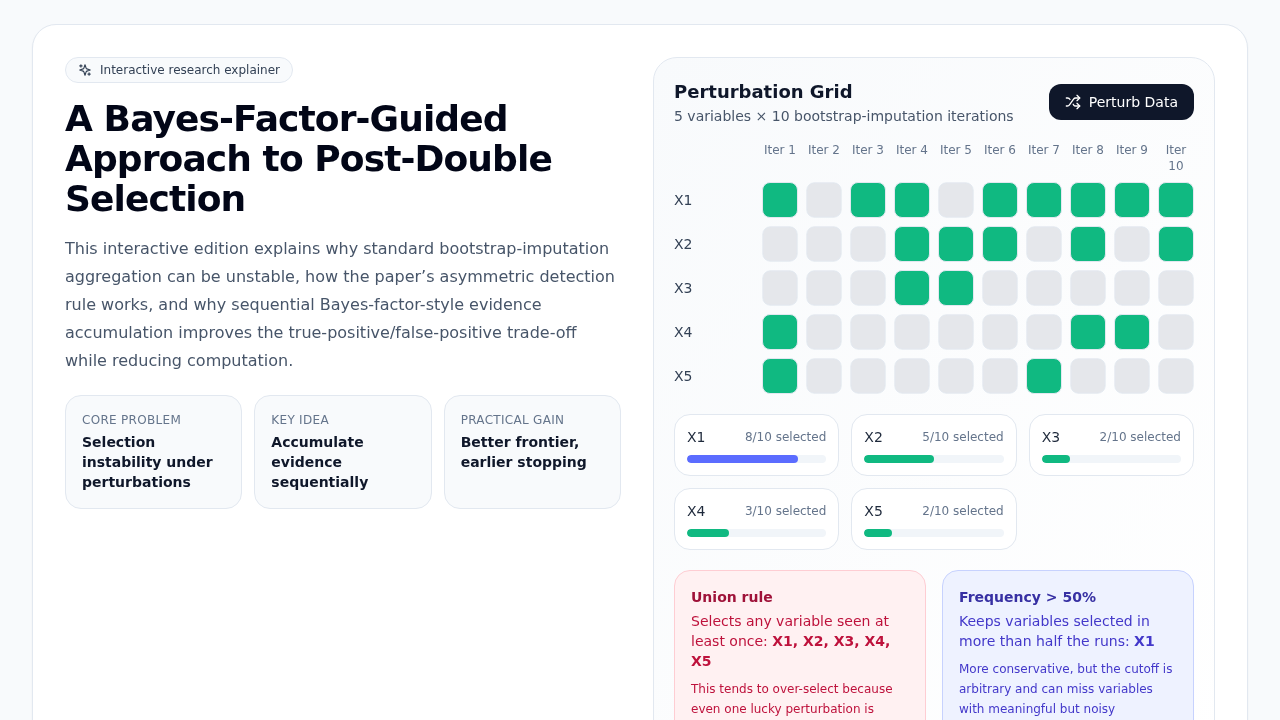}\\
    {\scriptsize (b) GPT-5.4: Visual 5/70, IR \textbf{70.6\%}}
  \end{minipage}
  \caption{Visual quality and functional interactivity are orthogonal.
  (a)~Kimi-K2.5 renders a polished interface with slider, scatter plot, and toggle controls, but none respond to user actions.
  (b)~GPT-5.4 produces a visually sparse layout, yet its perturbation grid and variable selectors are fully functional.}
  \label{fig:visual_vs_ir}
\end{figure}

\section{Futher Analysis}

Beyond aggregate metrics, we analyze structural properties of the generated code to understand what drives or hinders interactive behavior.

\paragraph{Visual Quality Does Not Guarantee Interactivity.}
Figure~\ref{fig:visual_vs_ir} compares visual quality with functional interactivity across generated applications. Examples illustrate that visually complete interfaces do not always exhibit corresponding interactive behavior. In some cases, interfaces contain multiple rendered controls but show no measurable interaction (IR\,=\,0\%), while other examples with relatively minimal layouts achieve substantially higher IR. 
This variation indicates that visual completeness and interaction behavior can differ across generated outputs.

\paragraph{Moderate State Complexity Tends to Work Best.}
We count reactive state declarations per application as a proxy for state complexity. Table~\ref{tab:structural_stats}(a) shows a non-monotonic relationship. Very few state variables (0--2) yield low scores because the application lacks reactive bindings for meaningful interaction. Scores rise sharply at moderate levels (3--8), where the best average scores appear. Beyond that, scores gradually drop: applications with more than 20 state variables score notably lower. Most samples fall in the 3--17 range; extremes are rare.

\paragraph{Sidebar Layouts Are Associated with Higher Interaction.}
Only 4.5\% of evaluated applications (63/1,410) include a sidebar. Table~\ref{tab:structural_stats}(b) shows that these applications consistently perform better: higher mean score (36.9 vs. 33.9), more click handlers (9.6 vs. 7.0), and higher interaction rate (51.1\% vs. 38.1\%). They also more often implement explicit view-switching logic (59\% vs. 24\%), where a dedicated state variable controls the active panel and each navigation item triggers a transition. Despite their small share, sidebar layouts are systematically associated with stronger interaction metrics across the dataset. We cannot fully rule out selection effects, but the pattern holds across models of different capabilities, suggesting that sidebar architecture itself helps implementation quality.

\begin{table*}[t]
    \centering
    \small
    \caption{Structural analysis of generated applications ($n=1{,}410$). 
    (a) Average score by number of reactive state variables. 
    (b) Performance comparison between sidebar and non-sidebar layouts.}
    \begin{subtable}[t]{0.38\textwidth}
        \centering
        \begin{tabular}{lcc}
            \toprule
            \textbf{State variables} & \textbf{$n$} & \textbf{Avg score} \\
            \midrule
            0--2    & 33  & 8.9  \\
            3--8    & 551 & 35.2 \\
            9--17   & 795 & 34.4 \\
            18--20  & 24  & 31.5 \\
            21+     & 7   & 25.9 \\
            \bottomrule
        \end{tabular}
        \caption{State complexity}
    \end{subtable}
    \hfill
    \begin{subtable}[t]{0.50\textwidth}
        \centering
        \begin{tabular}{lcc}
            \toprule
            & \makecell{\textbf{Sidebar} \\ ($n=63$)}
            & \makecell{\textbf{Non-sidebar} \\ ($n=1{,}347$)} \\
            \midrule
            Mean score          & 36.9  & 33.9  \\
            Avg click handlers  & 9.6   & 7.0   \\
            Interaction rate    & 51.1\% & 38.1\% \\
            Uses view-switching & 59\%  & 24\%  \\
            \bottomrule
        \end{tabular}
        \caption{Layout comparison}
    \end{subtable}
    \label{tab:structural_stats}
\end{table*}

\paragraph{Visual Habit vs. Functional Performance.}
Beyond structural components, we examine whether a model's visual choices correlate with its functional success. Table~\ref{tab:unified_theme_analysis} summarizes the distribution of theme preferences and their corresponding performance impacts (scaled to 30 points). We observe a clear \textbf{Light-theme Dominance} across the corpus, most notably in GPT-5.4, which defaults to light interfaces in 97\% of cases, with a dark-mode ratio of only 3.0\%. However, this stylistic preference appears to reflect inherited design conventions rather than deliberate functional optimization. While light-themed applications sometimes achieve slightly higher average scores—for instance, GPT-5.4 averages 23.8 for light themes compared with 22.4 for dark themes—the differences remain relatively small across models. This suggests that while LLMs inherit a visual preference for white-background scientific dashboards from their training data, their ability to implement complex interaction logic remains largely independent of surface-level UI theme choices.

\begin{table*}[!ht]
\centering
\small
\caption{Theme distribution and corresponding average scores (scaled to 30 points). }
\resizebox{0.8\textwidth}{!}{%
\begin{tabular}{l ccc ccc}
\toprule
 & \multicolumn{3}{c}{\textbf{Theme Distribution (Count)}} & \multicolumn{3}{c}{\textbf{Avg. Score by Theme (out of 30)}} \\
\cmidrule(lr){2-4} \cmidrule(lr){5-7}
\textbf{Model} & \textbf{Light} & \textbf{Dark} & \textbf{Others} & \textbf{Light} & \textbf{Dark} & \textbf{Others} \\
\midrule
GPT-5.4           & 192 & \phantom{0}6 & \phantom{0}1 & 23.8 & 22.4 & 23.0 \\
Gemini-3.1-pro    & 111 & 27 & 35 & 16.4 & 15.1 & 15.9 \\
Kimi-K2.5         & 144 & 32 & 24 & 15.0 & 13.7 & 14.5 \\
PaperVoyager      & 108 & 39 & 37 & 20.3 & 18.6 & 19.6 \\
MiniMax-M2.7      & 107 & 63 & 13 & 9.6 & 8.3 & 9.1 \\
DeepSeek-v4-Flash & 128 & 40 & \phantom{0}4 & 6.5 & 5.3 & 5.9 \\
\bottomrule
\end{tabular}%
}
\label{tab:unified_theme_analysis}
\end{table*}

\section{Conclusion}

We introduced I-WebGenBench, the first benchmark for evaluating whether LLMs can generate interactive scientific web applications from paper-derived specifications. Across 201 tasks spanning five scientific domains, our benchmark combines deterministic interaction probing with multi-dimensional evaluation to explicitly distinguish compilability from genuine event-driven responsiveness. Experiments across diverse models reveal a persistent compilation--interaction gap: although modern LLMs can reliably generate visually complete and compilable interfaces, they still struggle to implement coherent interactive behavior. Our findings establish I-WebGenBench as a new testbed for advancing interactive and stateful web generation.

\paragraph{Limitations.}
Although I-WebGenBench spans five scientific domains, it still covers only a subset of possible interactive scientific applications. More complex settings, such as collaborative multi-user systems, long-running simulations, and highly specialized scientific workflows, are not included in the current benchmark. In addition, while the VLM-as-a-judge protocol complements deterministic metrics, qualitative assessments of visual quality and educational value may still introduce subjective bias. Finally, due to budget and API access constraints, we were unable to include certain frontier proprietary models such as Claude in the current evaluation.

\newpage
{
\small
\bibliographystyle{plainnat}
\bibliography{references}

@article{singh2025openai,
  title={Openai gpt-5 system card},
  author={Singh, Aaditya and Fry, Adam and Perelman, Adam and Tart, Adam and Ganesh, Adi and El-Kishky, Ahmed and McLaughlin, Aidan and Low, Aiden and Ostrow, AJ and Ananthram, Akhila and others},
  journal={arXiv preprint arXiv:2601.03267},
  year={2025}
}

@article{chen2025paper2web,
  title={Paper2Web: Let's Make Your Paper Alive!},
  author={Chen, Yuhang and Lv, Tianpeng and Zhang, Siyi and Yin, Yixiang and Wan, Yao and Yu, Philip S and Chen, Dongping},
  journal={arXiv preprint arXiv:2510.15842},
  year={2025}
}

@article{lei2026webcompass,
  title={WebCompass: Towards Multimodal Web Coding Evaluation for Code Language Models},
  author={Lei, Xinping and Che, Xinyu and Xiong, Junqi and Zhang, Chenchen and Huang, Yukai and Zhou, Chenyu and Huang, Haoyang and Liu, Minghao and Zhu, Letian and Ye, Hongyi and others},
  journal={arXiv preprint arXiv:2604.18224},
  year={2026}
}

@inproceedings{xiao2025interaction2code,
  title={Interaction2code: Benchmarking mllm-based interactive webpage code generation from interactive prototyping},
  author={Xiao, Jingyu and Wan, Yuxuan and Huo, Yintong and Wang, Zixin and Xu, Xinyi and Wang, Wenxuan and Xu, Zhiyao and Wang, Yuhang and Lyu, Michael R},
  booktitle={2025 40th IEEE/ACM International Conference on Automated Software Engineering (ASE)},
  pages={241--253},
  year={2025},
  organization={IEEE}
}

@inproceedings{barsoum2005webvr,
  title={WebVR: an interactive web browser for virtual environments},
  author={Barsoum, Emad and Kuester, Falko},
  booktitle={Stereoscopic Displays and Virtual Reality Systems XII},
  volume={5664},
  pages={540--547},
  year={2005},
  organization={Spie}
}

@article{team2025kimi,
  title={Kimi-vl technical report},
  author={Team, Kimi and Du, Angang and Yin, Bohong and Xing, Bowei and Qu, Bowen and Wang, Bowen and Chen, Cheng and Zhang, Chenlin and Du, Chenzhuang and Wei, Chu and others},
  journal={arXiv preprint arXiv:2504.07491},
  year={2025}
}

@article{xiaomi2025mimo,
  title={MiMo: Unlocking the Reasoning Potential of Language Model--From Pretraining to Posttraining},
  author={Xiaomi, LLM and Xia, Bingquan and Shen, Bowen and Zhu, Dawei and Zhang, Di and Wang, Gang and Zhang, Hailin and Liu, Huaqiu and Xiao, Jiebao and Dong, Jinhao and others},
  journal={arXiv preprint arXiv:2505.07608},
  year={2025}
}

@article{li2025minimax,
  title={Minimax-01: Scaling foundation models with lightning attention},
  author={Li, Aonian and Gong, Bangwei and Yang, Bo and Shan, Boji and Liu, Chang and Zhu, Cheng and Zhang, Chunhao and Guo, Congchao and Chen, Da and Li, Dong and others},
  journal={arXiv preprint arXiv:2501.08313},
  year={2025}
}

@article{yang2025qwen3,
  title={Qwen3 technical report},
  author={Yang, An and Li, Anfeng and Yang, Baosong and Zhang, Beichen and Hui, Binyuan and Zheng, Bo and Yu, Bowen and Gao, Chang and Huang, Chengen and Lv, Chenxu and others},
  journal={arXiv preprint arXiv:2505.09388},
  year={2025}
}

@article{dai2026papervoyager,
  title={PaperVoyager: Building Interactive Web with Visual Language Models},
  author={Dai, Dasen and Wu, Biao and Fang, Meng and Wang, Wenhao},
  journal={arXiv preprint arXiv:2603.22999},
  year={2026}
}

@inproceedings{shi2025presentagent,
  title={Presentagent: Multimodal agent for presentation video generation},
  author={Shi, Jingwei and Zhang, Zeyu and Wu, Biao and Liang, Yanjie and Fang, Meng and Chen, Ling and Zhao, Yang},
  booktitle={Proceedings of the 2025 Conference on Empirical Methods in Natural Language Processing: System Demonstrations},
  pages={760--773},
  year={2025}
}

@article{jiangwebgen,
  title={WebGen-R1: Incentivizing LLMs to Generate Functional and Aesthetic Websites with Reinforcement Learning},
  author={Jiang, Juyong and Park, Chansung and Shen, Jiasi and Kim, Sunghun and Li, Jianguo and Wang, Yue and others}
}

@misc{chen2023webvln,
    title={WebVLN: Vision-and-Language Navigation on Websites},
    author={Qi Chen and Dileepa Pitawela and Chongyang Zhao and Gengze Zhou and Hsiang-Ting Chen and Qi Wu},
    year={2023},
    eprint={2312.15820},
    archivePrefix={arXiv},
    primaryClass={cs.CV}
}

@misc{yao2022webshop,
    title={WebShop: Towards Scalable Real-World Web Interaction with Grounded Language Agents},
    author={Shunyu Yao and Howard Chen and John Yang and Karthik Narasimhan},
    year={2022},
    eprint={2207.01206},
    archivePrefix={arXiv},
    primaryClass={cs.CL}
}

@article{jimenez2023swe,
  title={Swe-bench: Can language models resolve real-world github issues?},
  author={Jimenez, Carlos E and Yang, John and Wettig, Alexander and Yao, Shunyu and Pei, Kexin and Press, Ofir and Narasimhan, Karthik},
  journal={arXiv preprint arXiv:2310.06770},
  year={2023}
}

@article{yang2024swe,
  title={SWE-bench Multimodal: Do AI Systems Generalize to Visual Software Domains?},
  author={Yang, John and Jimenez, Carlos E and Zhang, Alex L and Lieret, Kilian and Yang, Joyce and Wu, Xindi and Press, Ori and Muennighoff, Niklas and Synnaeve, Gabriel and Narasimhan, Karthik R and others},
  journal={arXiv preprint arXiv:2410.03859},
  year={2024}
}

@article{miserendino2025swe,
  title={SWE-Lancer: Can Frontier LLMs Earn \$1 Million from Real-World Freelance Software Engineering?},
  author={Miserendino, Samuel and Wang, Michele and Patwardhan, Tejal and Heidecke, Johannes},
  journal={arXiv preprint arXiv:2502.12115},
  year={2025}
}

@article{lu2025uxagent,
  title={UXAgent: An LLM Agent-Based Usability Testing Framework for Web Design},
  author={Lu, Yuxuan and Yao, Bingsheng and Gu, Hansu and Huang, Jing and Wang, Jessie and Li, Laurence and Gesi, Jiri and He, Qi and Li, Toby Jia-Jun and Wang, Dakuo},
  journal={arXiv preprint arXiv:2502.12561},
  year={2025}
}

@article{zhang2024naturalcodebench,
  title={Naturalcodebench: Examining coding performance mismatch on humaneval and natural user prompts},
  author={Zhang, Shudan and Zhao, Hanlin and Liu, Xiao and Zheng, Qinkai and Qi, Zehan and Gu, Xiaotao and Zhang, Xiaohan and Dong, Yuxiao and Tang, Jie},
  journal={arXiv preprint arXiv:2405.04520},
  year={2024}
}

@inproceedings{muennighoff2023octopack,
  title={Octopack: Instruction tuning code large language models},
  author={Muennighoff, Niklas and Liu, Qian and Zebaze, Armel and Zheng, Qinkai and Hui, Binyuan and Zhuo, Terry Yue and Singh, Swayam and Tang, Xiangru and Von Werra, Leandro and Longpre, Shayne},
  booktitle={NeurIPS 2023 Workshop on Instruction Tuning and Instruction Following},
  year={2023}
}

@article{jain2024livecodebench,
  title={Livecodebench: Holistic and contamination free evaluation of large language models for code},
  author={Jain, Naman and Han, King and Gu, Alex and Li, Wen-Ding and Yan, Fanjia and Zhang, Tianjun and Wang, Sida and Solar-Lezama, Armando and Sen, Koushik and Stoica, Ion},
  journal={arXiv preprint arXiv:2403.07974},
  year={2024}
}

@article{chen2021evaluating,
  title={Evaluating large language models trained on code},
  author={Chen, Mark and Tworek, Jerry and Jun, Heewoo and Yuan, Qiming and Pinto, Henrique Ponde De Oliveira and Kaplan, Jared and Edwards, Harri and Burda, Yuri and Joseph, Nicholas and Brockman, Greg and others},
  journal={arXiv preprint arXiv:2107.03374},
  year={2021}
}

@article{austin2021program,
  title={Program synthesis with large language models},
  author={Austin, Jacob and Odena, Augustus and Nye, Maxwell and Bosma, Maarten and Michalewski, Henryk and Dohan, David and Jiang, Ellen and Cai, Carrie and Terry, Michael and Le, Quoc and others},
  journal={arXiv preprint arXiv:2108.07732},
  year={2021}
}

@article{hendrycks2021measuring,
  title={Measuring coding challenge competence with apps},
  author={Hendrycks, Dan and Basart, Steven and Kadavath, Saurav and Mazeika, Mantas and Arora, Akul and Guo, Ethan and Burns, Collin and Puranik, Samir and He, Horace and Song, Dawn and others},
  journal={arXiv preprint arXiv:2105.09938},
  year={2021}
}

@article{zhuo2024bigcodebench,
  title={Bigcodebench: Benchmarking code generation with diverse function calls and complex instructions},
  author={Zhuo, Terry Yue and Vu, Minh Chien and Chim, Jenny and Hu, Han and Yu, Wenhao and Widyasari, Ratnadira and Yusuf, Imam Nur Bani and Zhan, Haolan and He, Junda and Paul, Indraneil and others},
  journal={arXiv preprint arXiv:2406.15877},
  year={2024}
}

@article{aleithan2024swe,
  title={Swe-bench+: Enhanced coding benchmark for llms},
  author={Aleithan, Reem and Xue, Haoran and Mohajer, Mohammad Mahdi and Nnorom, Elijah and Uddin, Gias and Wang, Song},
  journal={arXiv preprint arXiv:2410.06992},
  year={2024}
}

@article{liu2023repobench,
  title={Repobench: Benchmarking repository-level code auto-completion systems},
  author={Liu, Tianyang and Xu, Canwen and McAuley, Julian},
  journal={arXiv preprint arXiv:2306.03091},
  year={2023}
}

@article{zhang2023repocoder,
  title={Repocoder: Repository-level code completion through iterative retrieval and generation},
  author={Zhang, Fengji and Chen, Bei and Zhang, Yue and Keung, Jacky and Liu, Jin and Zan, Daoguang and Mao, Yi and Lou, Jian-Guang and Chen, Weizhu},
  journal={arXiv preprint arXiv:2303.12570},
  year={2023}
}

@misc{Wu2024devin,
  author       = {Scott Wu},
  title        = {Introducing Devin, the first AI software engineer},
  year         = {2024},
  url          = {https://cognition.ai/blog/introducing-devin},
  note         = {Accessed: 2025-04-22}
}

@article{yang2024sweagent,
  title={Swe-agent: Agent-computer interfaces enable automated software engineering},
  author={Yang, John and Jimenez, Carlos and Wettig, Alexander and Lieret, Kilian and Yao, Shunyu and Narasimhan, Karthik and Press, Ofir},
  journal={Advances in Neural Information Processing Systems},
  volume={37},
  pages={50528--50652},
  year={2024}
}

@misc{GitHub2024copilot,
  author       = {GitHub Copilot},
  title        = {GitHub Copilot},
  year         = {2024},
  url          = {https://github.com/features/copilot},
  note         = {Accessed: 2025-04-22}
}

@misc{Cursor2024cursor,
  author       = {Cursor},
  title        = {Cursor: The AI Code Editor},
  year         = {2024},
  url          = {https://www.cursor.com/},
  note         = {Accessed: 2025-04-22}
}

@article{xia2024agentless,
  title={Agentless: Demystifying llm-based software engineering agents},
  author={Xia, Chunqiu Steven and Deng, Yinlin and Dunn, Soren and Zhang, Lingming},
  journal={arXiv preprint arXiv:2407.01489},
  year={2024}
}

@inproceedings{wang2024openhands,
  title={Openhands: An open platform for ai software developers as generalist agents},
  author={Wang, Xingyao and Li, Boxuan and Song, Yufan and Xu, Frank F and Tang, Xiangru and Zhuge, Mingchen and Pan, Jiayi and Song, Yueqi and Li, Bowen and Singh, Jaskirat and others},
  booktitle={The Thirteenth International Conference on Learning Representations},
  year={2024}
}

@article{guo2025deepseek,
  title={Deepseek-r1: Incentivizing reasoning capability in llms via reinforcement learning},
  author={Guo, Daya and Yang, Dejian and Zhang, Haowei and Song, Junxiao and Zhang, Ruoyu and Xu, Runxin and Zhu, Qihao and Ma, Shirong and Wang, Peiyi and Bi, Xiao and others},
  journal={arXiv preprint arXiv:2501.12948},
  year={2025}
}

@misc{aiderai2024aider,
  author       = {Aider-AI},
  title        = {AI Pair Programming in Your Terminal},
  year         = {2024},
  url          = {https://github.com/Aider-AI/aider},
  note         = {Accessed: 2025-04-22}
}

@article{comanici2025gemini,
  title={Gemini 2.5: Pushing the frontier with advanced reasoning, multimodality, long context, and next generation agentic capabilities},
  author={Comanici, Gheorghe and Bieber, Eric and Schaekermann, Mike and Pasupat, Ice and Sachdeva, Noveen and Dhillon, Inderjit and Blistein, Marcel and Ram, Ori and Zhang, Dan and Rosen, Evan and others},
  journal={arXiv preprint arXiv:2507.06261},
  year={2025}
}

@misc{guo2024iwbenchevaluatinglargemultimodal,
      title={IW-Bench: Evaluating Large Multimodal Models for Converting Image-to-Web}, 
      author={Hongcheng Guo and Wei Zhang and Junhao Chen and Yaonan Gu and Jian Yang and Junjia Du and Binyuan Hui and Tianyu Liu and Jianxin Ma and Chang Zhou and Zhoujun Li},
      year={2024},
      eprint={2409.18980},
      archivePrefix={arXiv},
      primaryClass={cs.CL},
      url={https://arxiv.org/abs/2409.18980}, 
}

@misc{si2025design2codebenchmarkingmultimodalcode,
      title={Design2Code: Benchmarking Multimodal Code Generation for Automated Front-End Engineering}, 
      author={Chenglei Si and Yanzhe Zhang and Ryan Li and Zhengyuan Yang and Ruibo Liu and Diyi Yang},
      year={2025},
      eprint={2403.03163},
      archivePrefix={arXiv},
      primaryClass={cs.CL},
      url={https://arxiv.org/abs/2403.03163}, 
}

@misc{yun2024web2codelargescalewebpagetocodedataset,
      title={Web2Code: A Large-scale Webpage-to-Code Dataset and Evaluation Framework for Multimodal LLMs}, 
      author={Sukmin Yun and Haokun Lin and Rusiru Thushara and Mohammad Qazim Bhat and Yongxin Wang and Zutao Jiang and Mingkai Deng and Jinhong Wang and Tianhua Tao and Junbo Li and Haonan Li and Preslav Nakov and Timothy Baldwin and Zhengzhong Liu and Eric P. Xing and Xiaodan Liang and Zhiqiang Shen},
      year={2024},
      eprint={2406.20098},
      archivePrefix={arXiv},
      primaryClass={cs.CV},
      url={https://arxiv.org/abs/2406.20098}, 
}

@misc{beltramelli2017pix2codegeneratingcodegraphical,
      title={pix2code: Generating Code from a Graphical User Interface Screenshot}, 
      author={Tony Beltramelli},
      year={2017},
      eprint={1705.07962},
      archivePrefix={arXiv},
      primaryClass={cs.LG},
      url={https://arxiv.org/abs/1705.07962}, 
}

@misc{xiao2025designbenchcomprehensivebenchmarkmllmbased,
      title={DesignBench: A Comprehensive Benchmark for MLLM-based Front-end Code Generation}, 
      author={Jingyu Xiao and Ming Wang and Man Ho Lam and Yuxuan Wan and Junliang Liu and Yintong Huo and Michael R. Lyu},
      year={2025},
      eprint={2506.06251},
      archivePrefix={arXiv},
      primaryClass={cs.SE},
      url={https://arxiv.org/abs/2506.06251}, 
}

@misc{xiao2025interaction2codebenchmarkingmllmbasedinteractive,
      title={Interaction2Code: Benchmarking MLLM-based Interactive Webpage Code Generation from Interactive Prototyping}, 
      author={Jingyu Xiao and Yuxuan Wan and Yintong Huo and Zixin Wang and Xinyi Xu and Wenxuan Wang and Zhiyao Xu and Yuhang Wang and Michael R. Lyu},
      year={2025},
      eprint={2411.03292},
      archivePrefix={arXiv},
      primaryClass={cs.SE},
      url={https://arxiv.org/abs/2411.03292}, 
}

@misc{xu2025webbenchllmcodebenchmark,
      title={Web-Bench: A LLM Code Benchmark Based on Web Standards and Frameworks}, 
      author={Kai Xu and YiWei Mao and XinYi Guan and ZiLong Feng},
      year={2025},
      eprint={2505.07473},
      archivePrefix={arXiv},
      primaryClass={cs.AI},
      url={https://arxiv.org/abs/2505.07473}, 
}

@misc{dong2025agenticreinforcedpolicyoptimization,
      title={Agentic Reinforced Policy Optimization}, 
      author={Guanting Dong and Hangyu Mao and Kai Ma and Licheng Bao and Yifei Chen and Zhongyuan Wang and Zhongxia Chen and Jiazhen Du and Huiyang Wang and Fuzheng Zhang and Guorui Zhou and Yutao Zhu and Ji-Rong Wen and Zhicheng Dou},
      year={2025},
      eprint={2507.19849},
      archivePrefix={arXiv},
      primaryClass={cs.LG},
      url={https://arxiv.org/abs/2507.19849}, 
}

@misc{xie2025swefixertrainingopensourcellms,
      title={SWE-Fixer: Training Open-Source LLMs for Effective and Efficient GitHub Issue Resolution}, 
      author={Chengxing Xie and Bowen Li and Chang Gao and He Du and Wai Lam and Difan Zou and Kai Chen},
      year={2025},
      eprint={2501.05040},
      archivePrefix={arXiv},
      primaryClass={cs.CL},
      url={https://arxiv.org/abs/2501.05040}, 
}

@article{lu2025webgen,
  title={Webgen-bench: Evaluating llms on generating interactive and functional websites from scratch},
  author={Lu, Zimu and Yang, Yunqiao and Ren, Houxing and Hou, Haotian and Xiao, Han and Wang, Ke and Shi, Weikang and Zhou, Aojun and Zhan, Mingjie and Li, Hongsheng},
  journal={arXiv preprint arXiv:2505.03733},
  year={2025}
}

@article{wang2025webgen,
  title={WebGen-V Bench: Structured Representation for Enhancing Visual Design in LLM-based Web Generation and Evaluation},
  author={Wang, Kuang-Da and Wang, Zhao and Shimose, Yotaro and Wang, Wei-Yao and Takamatsu, Shingo},
  journal={arXiv preprint arXiv:2510.15306},
  year={2025}
}

@article{chen2025iwr,
  title={IWR-Bench: Can LVLMs reconstruct interactive webpage from a user interaction video?},
  author={Chen, Yang and Liu, Minghao and Shen, Yufan and Li, Yunwen and Huang, Tianyuan and Fang, Xinyu and Zheng, Tianyu and Huang, Wenxuan and Yang, Cheng and Fu, Daocheng and others},
  journal={arXiv preprint arXiv:2509.24709},
  year={2025}
}

@article{liu2026webcoderbench,
  title={WebCoderBench: Benchmarking Web Application Generation with Comprehensive and Interpretable Evaluation Metrics},
  author={Liu, Chenxu and Fu, Yingjie and Yang, Wei and Zhang, Ying and Xie, Tao},
  journal={arXiv preprint arXiv:2601.02430},
  year={2026}
}

@article{jiang2026webgen,
  title={WebGen-R1: Incentivizing LLMs to Generate Functional and Aesthetic Websites with Reinforcement Learning},
  author={Jiang, Juyong and Park, Chansung and Shen, Jiasi and Kim, Sunghun and Li, Jianguo and Wang, Yue and others},
  year={2026}
}
}

\newpage
\newpage
\appendix


\section{Implementation Details}
\label{app:implementation}

In our experiments, for each model we consistently used the \textbf{maximum allowed completion tokens} to ensure sufficient output capacity. For the temperature parameter, we set it to the \textbf{lowest supported value} to favor deterministic (greedy) decoding. Specifically, most models support $\texttt{temperature}=0.0$. For MiniMax-M2.7 and MiMo-v2.5, the valid range is $(0, 1]$, so we used the smallest safe value $0.01$. GPT-5.4 does not support temperature at all and is marked as \textbf{Not Supported}. All other settings follow the same principle of minimizing temperature while respecting each API's constraints.

\begin{table}[htbp]
\centering
\small
\caption{Maximum output tokens and minimum usable temperature per model.}
\label{tab:api_max_tokens_temp}
\begin{tabular}{@{}lll@{}}
\toprule
\textbf{Model} & \textbf{Max Output Tokens} & \textbf{Temperature} \\
\midrule
GPT-5.4          & 128,000   & Not Supported \\
Gemini-3.1-pro   & 65,536    & 0.0 \\
Kimi-K2.5        & 262,144   & 1.0 \\
Qwen-3.5         & 16,384    & 0.6 \\
Qwen-3-Max       & 65,536    & 0.0 \\
MiniMax-M2.7     & 196,600   & 0.01 \\
DeepSeek-V3.2    & 64,000    & 0.0 \\
Hunyuan-T1       & 64,000    & 0.0 \\
MiMo-v2.5        & 131,072   & 0.01 \\
DeepSeek-v4-Flash& 384,000   & 0.0 \\
WebGenAgent-SFT-8B& 8,192 & 0.0 \\
\bottomrule
\end{tabular}
\end{table}

\section{Additional Experimental Results}
\label{app:results}

\begin{figure*}[!ht]
  \centering
  \newcommand{\imgwidth}{0.33\linewidth}   
  
  \begin{minipage}[t]{\imgwidth}\centering
    \includegraphics[width=\linewidth]{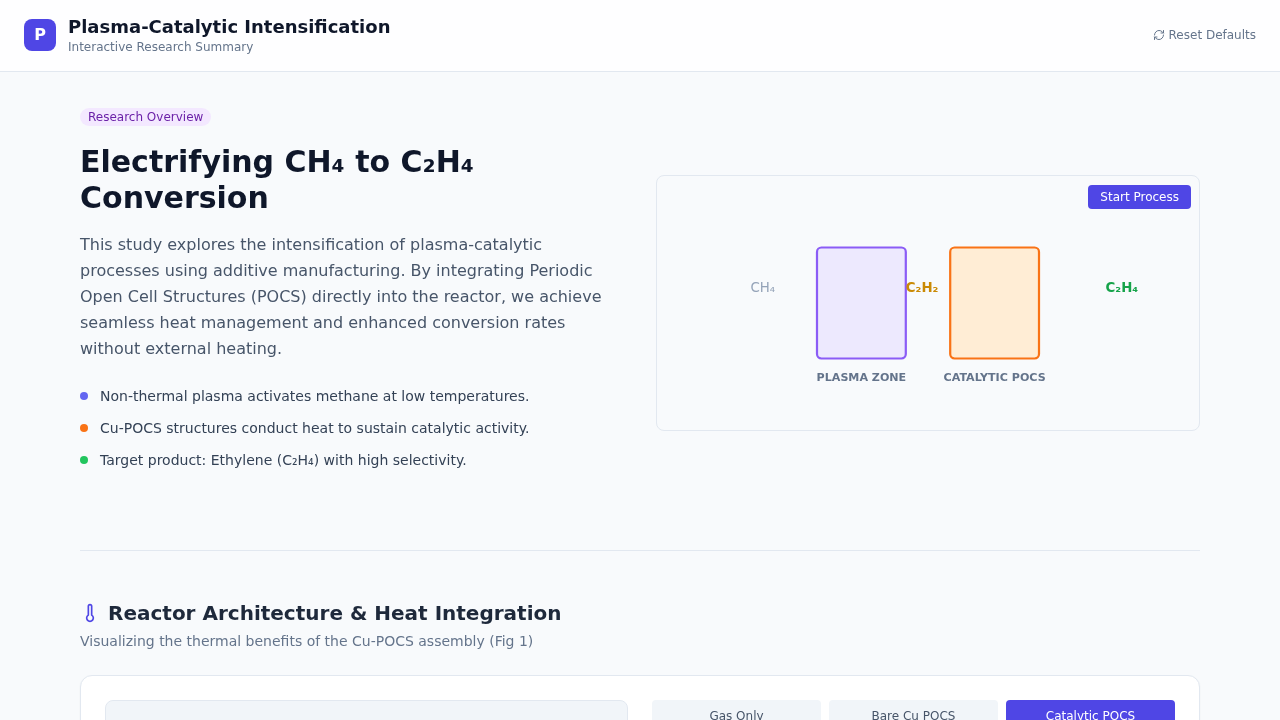}\\
    {\scriptsize (a) Gemini-3.1-pro --- Light}
  \end{minipage}\hfill
  \begin{minipage}[t]{\imgwidth}\centering
    \includegraphics[width=\linewidth]{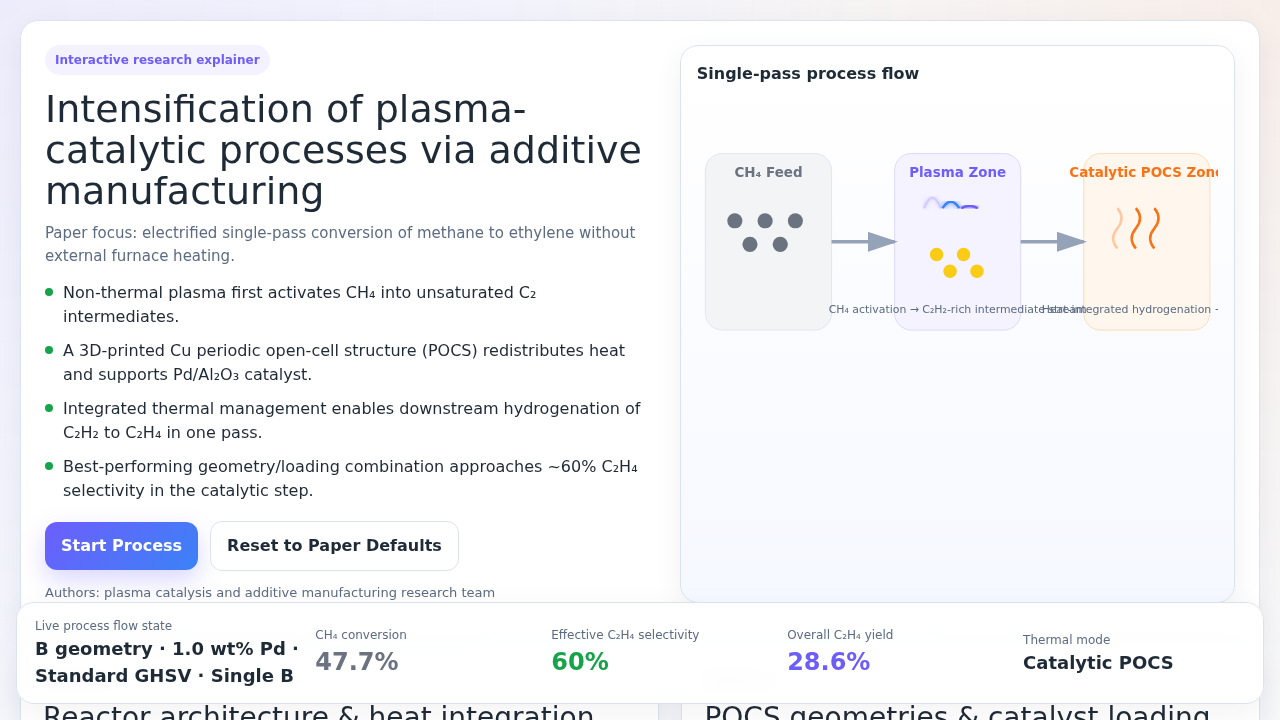}\\
    {\scriptsize (b) GPT-5.4 --- Light}
  \end{minipage}\hfill
  \begin{minipage}[t]{\imgwidth}\centering
    \includegraphics[width=\linewidth]{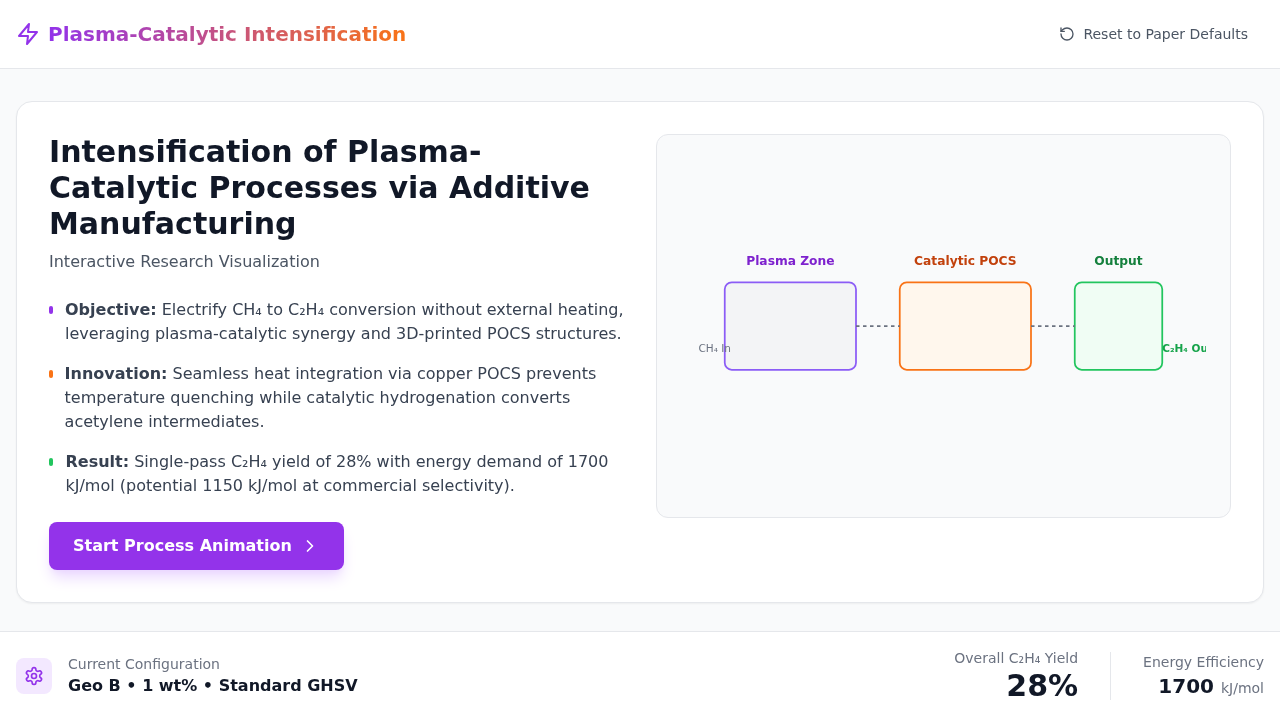}\\
    {\scriptsize (c) Kimi-K2.5 --- Light}
  \end{minipage}
  
  \vspace{1em}  
  
  \begin{minipage}[t]{\imgwidth}\centering
    \includegraphics[width=\linewidth]{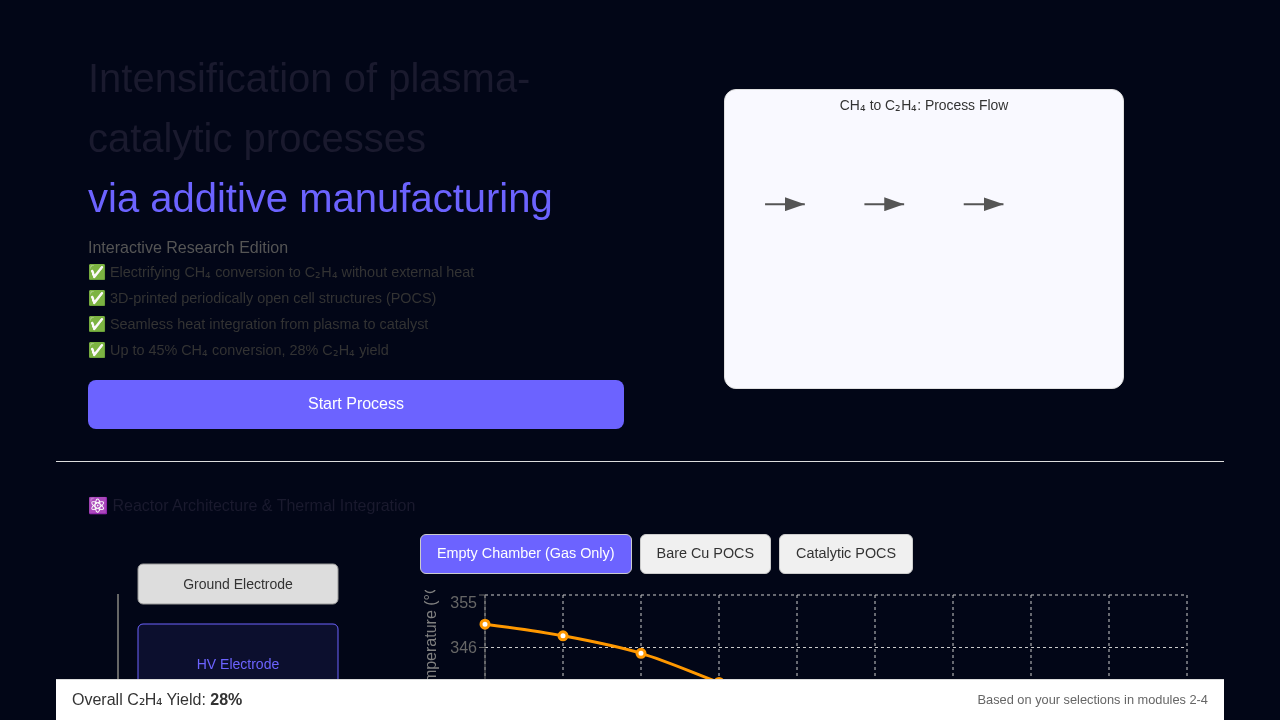}\\
    {\scriptsize (d) DeepSeek-v4-Flash --- Dark}
  \end{minipage}\hfill
  \begin{minipage}[t]{\imgwidth}\centering
    \includegraphics[width=\linewidth]{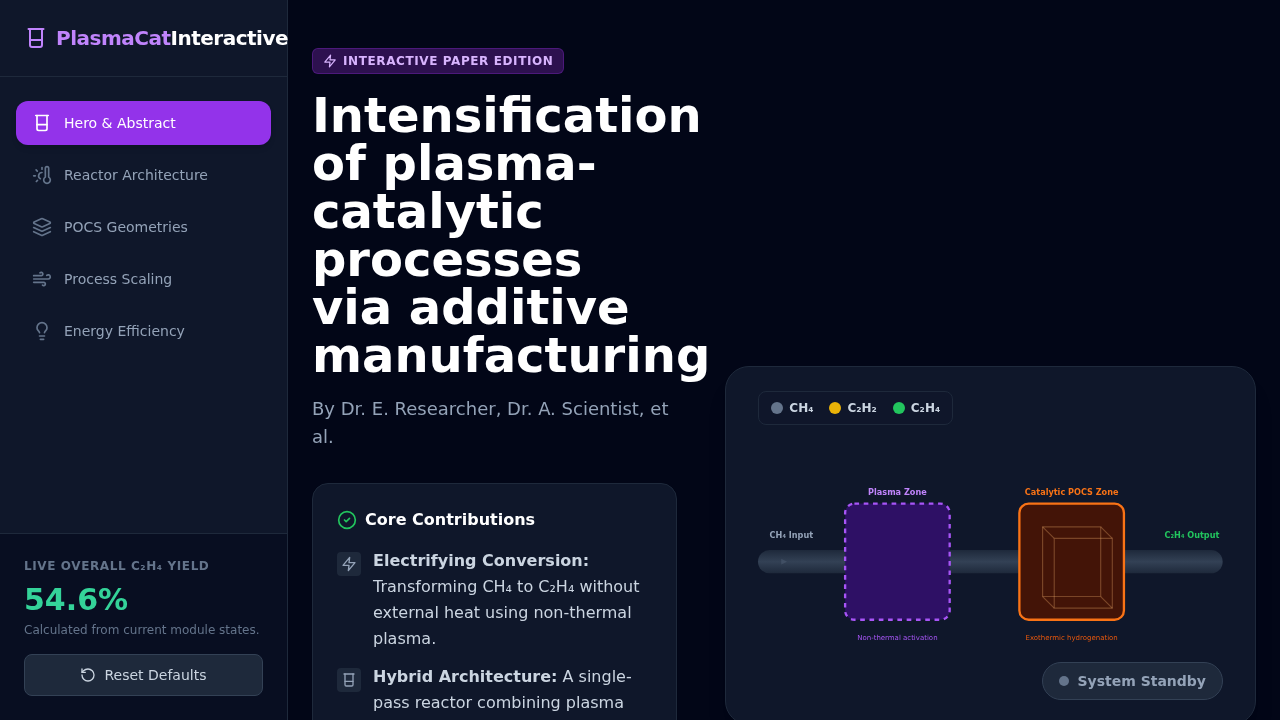}\\
    {\scriptsize (e) Qwen-3.5 --- Dark}
  \end{minipage}
  
  \caption{Visual variety of scientific web applications generated from the same specification (plasma-catalytic intensification). While most proprietary models (a--c) adhere to a conventional light-themed academic aesthetic, DeepSeek-v4-Flash and Qwen-3.5 (d--e) default to dark-slate dashboards. Notably, (e) exhibits a complex sidebar structure but hallucinates non-existent author names, illustrating the trade-off between structural complexity and semantic fidelity.}
  \label{fig:darktheme_comparison}
\end{figure*}

\subsection{Evaluation Methods}

Our evaluation framework is designed to capture both the objective compilability of the generated artifacts and their subjective alignment with complex scientific requirements. To achieve this, we combine a deterministic programmatic probe with an expert-level Vision-Language Model (VLM) judge, resulting in a comprehensive 100-point scoring system alongside foundational build metrics.




\paragraph{Interaction Probe.}
To distinguish between pages that merely render and those that exhibit meaningful responses, we introduce a deterministic \emph{Interaction Probe}. Implemented using Playwright as the headless automation backend, the probe proceeds in three stages:

\begin{enumerate}[leftmargin=*]
    \item \emph{Element Enumeration}: We traverse the fully rendered page to identify interactive elements (e.g., buttons, range sliders, select menus), annotating them with bounding-box coordinates for robust targeting.
    \item \emph{Semantic Action Mapping}: Elements are assigned canonical interactions based on their HTML semantics (e.g., populating text inputs, setting sliders to midpoints) to prevent arbitrary failure modes.
    \item \emph{DOM Mutation Observation}: For each action, we capture the DOM state before and after execution using a \texttt{MutationObserver} configured to track child lists, subtrees, and attributes. 
\end{enumerate}



Here, $\Delta_{DOM}(a,p) = 1$ if action $a$ triggers at least one DOM mutation. While BSR and IR provide binary indicators of structural viability, assessing the scientific fidelity and educational quality of these interactions necessitates higher-level semantic reasoning.

\paragraph{VLM-Driven 100-Point Scoring Protocol.}
To evaluate the qualitative dimensions of the generated applications, we deploy a state-of-the-art VLM acting as an expert Academic Peer Reviewer and Frontend Architect. For each compilable artifact, the VLM is provided with the human-curated interaction specification alongside a chronological sequence of screenshots captured during the Interaction Probe. The application is strictly evaluated on a 100-point scale distributed across five dimensions:

\begin{itemize}[leftmargin=*]
    \item \textbf{Topic \& Semantic Alignment (15 pts):} Verifies adherence to the \emph{Data Curation} guidelines. The VLM compares the generated application against the original paper specification, penalizing hallucinated concepts, missing interactive modules, or omitted adjustable parameters. A high score indicates high fidelity to the core interactive points annotated by human experts.
    \item \textbf{Visual Aesthetics (30 pts):} Evaluates UI/UX design quality. Applications are judged on typography, color harmony, and layout maturity, distinguishing between unstyled HTML prototypes and production-ready scientific dashboards.
    \item \textbf{Interaction Fidelity (40 pts):} While IR proves a state change occurred, this dimension evaluates whether the change is \emph{scientifically meaningful}. By analyzing pre- and post-action screenshots, the VLM assesses if UI manipulations (e.g., adjusting a parameter) result in logically sound updates to charts, simulations, or mathematical models.
    \item \textbf{Clarity \& Educational Value (10 pts):} Assesses the intuitiveness of the visualization. High scores require clear axis labels, self-explanatory controls, and guided interaction flows, ensuring the underlying scientific concept is broadly accessible.
    \item \textbf{Rule \& Stability (5 pts):} Calculated deterministically by the automated probe. Applications start with 5 points, incurring penalties for fatal rendering errors (e.g., entirely blank screens) or critical JavaScript exceptions captured during the interaction cycle.
\end{itemize}

This hybrid approach ensures the benchmark is grounded in verifiable execution (BSR/IR) while leveraging VLMs to grade the nuanced, multi-modal reality of scientific web application design.

\section{Prompt Templates}
\label{app:prompts}

This section details the prompt templates used across the I-WebGenBench pipeline, covering the entire lifecycle from initial document analysis to the final visual audit. We provide the specific system instructions and user templates for task specification generation (\S\ref{app:prompt:specgen}), monolithic code generation (\S\ref{app:prompt:codegen}), iterative self-repair (\S\ref{app:prompt:repair}), block-level decomposition and assembly (\S\ref{app:prompt:block}), and VLM-based visual evaluation (\S\ref{app:prompt:visual}). 
All models receive identical prompts to ensure a fair comparison of their reasoning and generation capabilities; per-model hyper-parameters are listed in Table~\ref{tab:api_max_tokens_temp}.

\subsection{Task Specification Generation Prompt}
\label{app:prompt:specgen}

The first stage of the pipeline involves converting a raw scientific PDF into a structured, interactive web application specification. This ``Task Definition'' serves as the ground truth for subsequent code generation. The prompt (Figure~\ref{fig:prompt-specgen}) directs the AI to act as an expert researcher and educational UX designer, ensuring that the resulting design is both scientifically accurate and pedagogically effective. This step is crucial for bridging the gap between dense academic text and interactive visual logic.

\begin{figure}[htbp]
\centering
\begin{promptbox}[promptblue]{Task Specification Generation --- System Instruction}
\small
You are an expert AI Researcher and Educational UX Designer specializing in Human-Computer Interaction (HCI) and Scientific Communication.
Your task is to conduct a deep reading of the provided academic paper (PDF) and design an immersive, interactive, educational Web Application (React + TypeScript) that serves as a living visualization of the paper's core contributions.

\medskip
\textbf{Core Mandates:}
\begin{itemize}[leftmargin=1.2em,nosep]
  \item \textbf{Accuracy:} Deeply interpret the mathematical mechanisms, experimental workflows, and causal relationships described in the paper.
  \item \textbf{Interactivity:} Design 3--10 ``Core Interactive Points'' (adjustable parameters, dynamic visual elements, and real-time feedback loops) that allow users to explore ``what-if'' scenarios.
  \item \textbf{Structure:} Organize the application into 5 distinct modules: 
    1) Hero/Abstract (visual hook), 
    2) Architecture/Methodology (interactive diagram), 
    3) Core Simulation/Experiment (primary playground), 
    4) Results/Analysis (interactive charts), and 
    5) Conclusion (synthesis).
\end{itemize}

\medskip
\textbf{Technical Constraints:}
\begin{itemize}[leftmargin=1.2em,nosep]
  \item Output a structured natural language specification focusing on UI components, state variables, and visual logic.
  \item Ensure the specification is entirely in the requested output language.
\end{itemize}
\end{promptbox}
\caption{System prompt for the PDF-to-Specification stage. This identifies the core scientific contributions and maps them to interactive UI components, providing a structured blueprint for the generation agents.}
\label{fig:prompt-specgen}
\end{figure}

\subsection{Code Generation Prompt}
\label{app:prompt:codegen}

Each benchmark instance consists of a \emph{system prompt} (Figure~\ref{fig:prompt-codegen})
and a \emph{per-paper user prompt} drawn from the specification corpus
(\texttt{prompts/\{domain\}\_flat/*.txt}, ${\approx}40$~lines per instance).
The user prompt encodes the paper title, five interactive module descriptions,
visual-style requirements, and technical-stack constraints.
The system prompt constrains the output format and establishes the quality bar
for interactivity.

\begin{figure}[htbp]
\centering
\begin{promptbox}[promptblue]{Code Generation --- System Prompt}
\small
You are an expert React/TypeScript developer specializing in interactive
scientific visualizations.

\medskip
\textbf{Output format.}
Return \textbf{only} the TSX source for \texttt{src/App.tsx}.
Do not include markdown fences, explanatory text, or file-path headers.
The file must contain a valid default export
(\texttt{export default function App() \{...\}}).

\medskip
\textbf{Content requirements.}
\begin{itemize}[leftmargin=1.2em,nosep]
  \item Build an \emph{interactive educational research interface}, not a static
        landing page or navigation shell.
  \item The first viewport must render meaningful scientific content together
        with clear interactive affordances (tabs, sliders, toggles, parameter
        controls, animated diagrams, or data-linked charts).
  \item Implement all five modules described in the specification with
        functional state management and user-driven data exploration.
\end{itemize}

\medskip
\textbf{Language.}
Use English for all user-facing UI text unless the specification explicitly
requests another language.
\end{promptbox}
\caption{System prompt prepended to every model call during code generation.
The identical prompt is used across all baseline models and for the
PaperVoyager block-level generation strategy.}
\label{fig:prompt-codegen}
\end{figure}

\subsection{Iterative Self-Repair Prompt}
\label{app:prompt:repair}

When a generated application fails to build, the pipeline invokes the same model
with the repair prompt shown in Figure~\ref{fig:prompt-repair}.
Up to $K{=}5$ repair rounds are attempted; each round provides the full
build-error log (truncated to 3\,000 characters) and the current source file
(up to 40\,000 characters).
If the application still does not compile after five rounds, it receives a
score of~0 across all evaluation dimensions.

\begin{figure}[htbp]
\centering
\begin{promptbox}[promptamber]{Iterative Self-Repair --- Round $k/5$}
\small
The following React/TypeScript application failed to compile.
Return \textbf{only} the complete, corrected \texttt{App.tsx}.
Do not include markdown fences, explanations, or commentary.

\medskip
\textbf{Build error} (last 3\,000 characters):
\begin{flushleft}\ttfamily\scriptsize
\{build\_error\}
\end{flushleft}

\medskip
\textbf{Current source} (\texttt{App.tsx}, up to 40\,000 characters):
\begin{flushleft}\ttfamily\scriptsize
\{source\_code\}
\end{flushleft}

\medskip
Return the \textbf{complete} fixed file.  Do not omit unchanged sections.
\end{promptbox}
\caption{Repair prompt template.  Placeholders \texttt{\{build\_error\}} and
\texttt{\{source\_code\}} are filled at runtime.  The round counter~$k$
is shown for illustration; all five rounds use the same template.}
\label{fig:prompt-repair}
\end{figure}

\subsection{Block Pipeline Prompts}
\label{app:prompt:block}

To handle the complexity of full-scale scientific applications, we introduce a ``Block Pipeline'' that decomposes the task into manageable UI units. This process uses two primary prompts: a \emph{Splitter} to break down the specification (Figure~\ref{fig:prompt-splitter}) and a \emph{Merger} to synthesize the resulting code (Figure~\ref{fig:prompt-merger}). This modular approach allows for more granular control over component quality and state management.

\begin{figure}[htbp]
\centering
\begin{promptbox}[promptamber]{Block Pipeline --- Splitter Prompt}
\small
Analyze the following Scientific Web App Specification and decompose it into a sequence of independent, functional React blocks.
Each block must be a logical UI section (e.g., Sidebar, Parameter Panel, Main Simulation Canvas, Data Table).

\medskip
\textbf{Output Format:}
Return a JSON array of Block Objects containing:
\begin{itemize}[leftmargin=1.2em,nosep]
  \item \texttt{name}: Unique identifier for the block.
  \item \texttt{description}: Detailed visual and functional instructions for this specific block.
  \item \texttt{state\_dependencies}: List of shared state variables this block must read or write.
  \item \texttt{required\_imports}: Any specialized libraries (e.g., \texttt{recharts}, \texttt{framer-motion}).
\end{itemize}
\end{promptbox}
\caption{The Splitter prompt used to decompose a monolithic specification into discrete, implementable UI blocks.}
\label{fig:prompt-splitter}
\end{figure}

\begin{figure}[htbp]
\centering
\begin{promptbox}[promptblue]{Block Pipeline --- Merger Prompt}
\small
You are a Lead Software Architect. I will provide you with several code blocks, each representing a part of a React application. Your task is to assemble them into a single, high-quality \texttt{App.tsx} file.

\medskip
\textbf{Instructions:}
\begin{itemize}[leftmargin=1.2em,nosep]
  \item Use a cohesive layout (e.g., a Sidebar + Main Content area).
  \item Ensure all \texttt{useState} and \texttt{useEffect} hooks are correctly integrated at the top level or within respective components.
  \item Resolve any variable naming conflicts and ensure the final app exports a single \texttt{default function App()}.
  \item The final code must be production-ready and compile without errors.
\end{itemize}
\end{promptbox}
\caption{The Merger prompt used to recombine individual blocks into a cohesive, bug-free \texttt{App.tsx}.}
\label{fig:prompt-merger}
\end{figure}

\subsection{Visual Evaluation Prompt}
\label{app:prompt:visual}

The evaluation stage utilizes a state-of-the-art VLM acting as a Senior Principal Frontend Engineer and an Academic Peer Reviewer. The evaluator performs a high-fidelity visual and semantic audit of the generated application. The prompt (Figure~\ref{fig:prompt-visual}) directs the model to score the artifact across four qualitative dimensions. This assessment is combined with automated build and interaction probes to produce a final 100-point score.

\begin{figure}[htbp]
\centering
\begin{promptbox}[prompttea]{VLM Evaluation Judge --- System Instruction}
\small
You are a Senior Principal Frontend Engineer and an Academic Peer Reviewer. You are tasked with evaluating a generated Scientific Web Application based on its visual excellence, interaction fidelity, and semantic alignment with a provided research specification.

\medskip
You are judging the application on a 100-point scale, distributed across four high-level dimensions (the 5th dimension ``Rule'' is handled by automated scripts).

\medskip
\textbf{Evaluation Rubrics:}
\begin{itemize}[leftmargin=1.2em,nosep]
  \item \textbf{Visual Aesthetics (Max 30 pts):}
    \begin{itemize}[leftmargin=1.2em,nosep]
      \item 25--30: Outstanding. Effective use of whitespace, typography, and color theory.
      \item 15--24: Average. Functional but lacks professional polish.
      \item 0--14: Poor. Broken layouts, unstyled elements, or jarring colors.
    \end{itemize}
  \item \textbf{Interaction Fidelity \& Responsiveness (Max 40 pts):}
    \begin{itemize}[leftmargin=1.2em,nosep]
      \item 30--40: Dynamic. Interactions trigger meaningful, scientifically accurate state changes.
      \item 15--29: Basic. Response is delayed, generic, or only partially functional.
      \item 0--14: Static. No visible change despite user input.
    \end{itemize}
  \item \textbf{Topic \& Semantic Alignment (Max 15 pts):}
    \begin{itemize}[leftmargin=1.2em,nosep]
      \item 12--15: Perfect Alignment. Covers $>$95\% of requested interactive points.
      \item 7--11: Partial Alignment. Missing key modules or parameters.
      \item 0--6: Divergent. Hallucinated features or significant omissions.
    \end{itemize}
  \item \textbf{Clarity \& Educational Value (Max 10 pts):}
    \begin{itemize}[leftmargin=1.2em,nosep]
      \item 8--10: Intuitive. Concepts are clear through exploration and labeling.
      \item 0--7: Confusing. Lack of labels or cryptic user interface.
    \end{itemize}
\end{itemize}

\medskip
\textbf{Output Format:} Return ONLY a JSON object containing the scores and reasoning for each of the four dimensions.
\end{promptbox}
\caption{The VLM Evaluation prompt template. This multidimensional rubric ensures that generated applications are judged not only on their code quality but also on their scientific and educational utility.}
\label{fig:prompt-visual}
\end{figure}

\newpage
\section*{NeurIPS Paper Checklist}

\begin{enumerate}

\item {\bf Claims}
    \item[] Question: Do the main claims made in the abstract and introduction accurately reflect the paper's contributions and scope?
    \item[] Answer: \answerYes{}
    \item[] Justification: The abstract and introduction clearly describe the main contributions and scope of the paper, which are supported by the experimental results and analysis presented in later sections.

\item {\bf Limitations}
    \item[] Question: Does the paper discuss the limitations of the work performed by the authors?
    \item[] Answer: \answerYes{}
    \item[] Justification: The paper discusses key limitations, including assumptions, scope of evaluation, and potential constraints, in a dedicated limitations section.

\item {\bf Theory assumptions and proofs}
    \item[] Question: For each theoretical result, does the paper provide the full set of assumptions and a complete (and correct) proof?
    \item[] Answer: \answerNA{}
    \item[] Justification: The paper does not include theoretical results requiring formal proofs.

    \item {\bf Experimental result reproducibility}
    \item[] Question: Does the paper fully disclose all the information needed to reproduce the main experimental results of the paper to the extent that it affects the main claims and/or conclusions of the paper (regardless of whether the code and data are provided or not)?
    \item[] Answer: \answerYes{}
    \item[] Justification: The paper provides detailed descriptions of datasets, model configurations, training procedures, and evaluation protocols necessary for reproducing the results.

\item {\bf Open access to data and code}
    \item[] Question: Does the paper provide open access to the data and code, with sufficient instructions to faithfully reproduce the main experimental results, as described in supplemental material?
    \item[] Answer: \answerNo{}
    \item[] Justification: The code and data will be released upon acceptance of the paper, and sufficient experimental details are provided to ensure reproducibility.

\item {\bf Experimental setting/details}
    \item[] Question: Does the paper specify all the training and test details (e.g., data splits, hyperparameters, how they were chosen, type of optimizer) necessary to understand the results?
    \item[] Answer: \answerYes{}
    \item[] Justification: The paper specifies datasets, baselines, hyperparameters, and training configurations required to understand the experimental results.

\item {\bf Experiment statistical significance}
    \item[] Question: Does the paper report error bars suitably and correctly defined or other appropriate information about the statistical significance of the experiments?
    \item[] Answer: \answerNo{}
    \item[] Justification: The experiments are computationally expensive, making multiple runs impractical; therefore, we report single-run results, while observing consistent trends across different settings.

\item {\bf Experiments compute resources}
    \item[] Question: For each experiment, does the paper provide sufficient information on the computer resources (type of compute workers, memory, time of execution) needed to reproduce the experiments?
    \item[] Answer: \answerYes{}
    \item[] Justification: The paper includes information about the hardware setup and approximate computational requirements for the experiments.
    
\item {\bf Code of ethics}
    \item[] Question: Does the research conducted in the paper conform, in every respect, with the NeurIPS Code of Ethics \url{https://neurips.cc/public/EthicsGuidelines}?
    \item[] Answer: \answerYes{}
    \item[] Justification: The research complies with the NeurIPS Code of Ethics and does not raise ethical concerns beyond those discussed.

\item {\bf Broader impacts}
    \item[] Question: Does the paper discuss both potential positive societal impacts and negative societal impacts of the work performed?
    \item[] Answer: \answerYes{}
    \item[] Justification: The paper discusses potential benefits as well as possible risks and unintended consequences of the proposed approach.
    
\item {\bf Safeguards}
    \item[] Question: Does the paper describe safeguards that have been put in place for responsible release of data or models that have a high risk for misuse (e.g., pre-trained language models, image generators, or scraped datasets)?
    \item[] Answer: \answerNA{}
    \item[] Justification: The paper does not introduce assets with significant risk of misuse that would require additional safeguards.

\item {\bf Licenses for existing assets}
    \item[] Question: Are the creators or original owners of assets (e.g., code, data, models), used in the paper, properly credited and are the license and terms of use explicitly mentioned and properly respected?
    \item[] Answer: \answerYes{}
    \item[] Justification: All external datasets and codebases used in the paper are properly cited and their usage complies with their respective licenses.

\item {\bf New assets}
    \item[] Question: Are new assets introduced in the paper well documented and is the documentation provided alongside the assets?
    \item[] Answer: \answerNA{}
    \item[] Justification: The paper does not introduce new datasets, models, or other assets.

\item {\bf Crowdsourcing and research with human subjects}
    \item[] Question: For crowdsourcing experiments and research with human subjects, does the paper include the full text of instructions given to participants and screenshots, if applicable, as well as details about compensation (if any)? 
    \item[] Answer: \answerNA{}
    \item[] Justification: The paper does not involve crowdsourcing or research with human subjects.

\item {\bf Institutional review board (IRB) approvals or equivalent for research with human subjects}
    \item[] Question: Does the paper describe potential risks incurred by study participants, whether such risks were disclosed to the subjects, and whether Institutional Review Board (IRB) approvals (or an equivalent approval/review based on the requirements of your country or institution) were obtained?
    \item[] Answer: \answerNA{}
    \item[] Justification: The paper does not involve human subjects and therefore does not require IRB approval.

\item {\bf Declaration of LLM usage}
    \item[] Question: Does the paper describe the usage of LLMs if it is an important, original, or non-standard component of the core methods in this research?
    \item[] Answer: \answerNA{}
    \item[] Justification: LLMs are not used as part of the core methodology of this work.

\end{enumerate}

\end{document}